\documentclass[lettersize,journal]{IEEEtran}
\usepackage{algorithm}
\usepackage{algpseudocode}
\usepackage{array}
\usepackage{svg}
\usepackage{graphicx}
\usepackage{subcaption} 
\usepackage{textcomp}
\usepackage{stfloats}
\usepackage{url}
\usepackage{verbatim}
\usepackage{cite}
\usepackage{picinpar}
\usepackage{amsmath,amsfonts}
\usepackage{flushend}
\usepackage[latin1]{inputenc}
\usepackage{colortbl}
\usepackage{soul}
\usepackage{multirow}
\usepackage{pifont}
\usepackage{color}
\usepackage{alltt}

\usepackage{enumerate}
\usepackage{siunitx}
\usepackage{breakurl}
\usepackage{epstopdf}
\usepackage{pbox}
\usepackage{xcolor}
\usepackage{booktabs}
\usepackage{gensymb}
\usepackage[hidelinks]{hyperref}

\usepackage{threeparttable}
\usepackage{amssymb} 
\usepackage{tikz}
\usepackage{listings}
\usepackage{authblk}
\begin{document}

\title{V2X-RECT: An Efficient V2X Trajectory Prediction Framework via Redundant Interaction Filtering and Tracking Error Correction}
\author{Xiangyan Kong, Xuecheng Wu, Xiongwei Zhao, Xiaodong Li, Yunyun Shi, Gang Wang,~\IEEEmembership{Senior Member,~IEEE}, Dingkang Yang, Yang Liu$^{\dagger}$, Hong Chen,~\IEEEmembership{Fellow,~IEEE}, Yulong Gao$^{\dagger}$
\thanks{This work was supported by the National Natural Science Foundation of China under Grant 62171163.} 
\thanks{Xiangyan Kong, Xiaodong Li, Gang Wang, and Yulong Gao are with the Communication Research Center, Harbin Institute of Technology, Harbin, 150001, China. (E-mail: 22b905071@stu.hit.edu.cn; lixd@stu.hit.edu.cn; gwang51@hit.edu.cn; ylgao@hit.edu.cn).}
\thanks{Xuecheng Wu and Yunyun Shi are with the School of Computer Science and Technology, Xi'an Jiaotong University, Xi'an, 710049, China. (E-mail: wuxc3@stu.xjtu.edu.cn; yunyunshi@stu.xjtu.edu.cn).}
\thanks{Xiongwei Zhao is with the School of Electronics and Information Engineering, Harbin Institute of Technology (Shenzhen), Shenzhen, 518055, China. (E-mail: xwzhao@stu.hit.edu.cn).}
\thanks{Dingkang Yang is with the College of Intelligent Robotics and Advanced Manufacturing, Fudan University, Shanghai, 200433, China. (E-mail: dkyang20@fudan.edu.cn).}
\thanks{Yang Liu and Hong Chen are with the College of Electronic and Information Engineering, Tongji University, Shanghai, 201804, China. (E-mail: yang\_liu@ieee.org; chenhong2019@tongji.edu.cn).}
\thanks{$\dagger$Corresponding authors: Yulong Gao \& Yang Liu.}
\thanks{Manuscript received XX, 2025; revised XX, 2025.}
}

\markboth{IEEE Transactions on Intelligent Transportation Systems}
{Shell \MakeLowercase{\textit{et al.}}: A Sample Article Using IEEEtran.cls for IEEE Journals}

\maketitle

\begin{abstract}
In the Connected-Automated vehicles, Vehicle-to-Everything (V2X) prediction can alleviate perception incompleteness caused by limited line of sight through fusing trajectory data from infrastructure and vehicles, which is crucial to traffic safety and efficiency. However, in dense traffic scenarios, frequent identity switching of targets hinders cross-view association and fusion. Meanwhile, multi-source information tends to generate redundant interactions during the encoding stage, and traditional vehicle-centric encoding leads to large amounts of repetitive historical trajectory feature encoding, degrading real-time inference performance. To address these challenges, we propose V2X-RECT, a trajectory prediction framework designed for high-density environments. It enhances data association consistency, reduces redundant interactions, and reuses historical information to enable more efficient and accurate prediction. Specifically, we design a multi-source identity matching and correction module that leverages multi-view spatiotemporal relationships to achieve stable and consistent target association, mitigating the adverse effects of mismatches on trajectory encoding and cross-view feature fusion. Then we introduce traffic signal-guided interaction module, encoding trend of traffic light changes as features and exploiting their role in constraining spatiotemporal passage rights to accurately filter key interacting vehicles, while capturing the dynamic impact of signal changes on interaction patterns. Furthermore, a local spatiotemporal coordinate encoding enables reusable features of historical trajectories and map, supporting parallel decoding and significantly improving inference efficiency. Extensive experimental results across V2X-Seq and V2X-Traj datasets demonstrate that our V2X-RECT achieves significant improvements across all the key performance indicators compared to the state-of-the-art methods, while also enhancing robustness and inference efficiency across diverse traffic densities.
\end{abstract}

\begin{IEEEkeywords}
Trajectory prediction, V2X, ID switch, Redundant interaction, Traffic signal, Feature reuse.
\end{IEEEkeywords}

\section{Introduction}

\IEEEPARstart{A}{utonomous} driving technology is widely regarded as a key pillar of future intelligent transportation systems. Accurately perceiving and predicting the future trajectories of traffic participants is fundamental to the safe operation of autonomous vehicles\cite{refe1*},\cite{refe2*},\cite{refe3*},\cite{refe2} and, at the same time, supports collaborative control and efficiency optimization in the transportation systems \cite{refe36*},\cite{refe38*}. However, realizing high accuracy trajectory prediction still faces numerous challenges. Existing single-vehicle systems\cite{refe8*},\cite{refe9*},\cite{refe10*} are constrained by limited line of sight, which is difficult to reliably obtain comprehensive trajectory information in complex urban environments. To address this limitation, collaborative prediction\cite{refe5*},\cite{refe6*},\cite{refe7*} by enabling data sharing and feature fusion among ego vehicle, infrastructures, and other vehicles, can partially mitigate the limitations caused by restricted visibility. This technology is expected to achieve more accurate spatiotemporal reasoning and a higher level of traffic intelligence. \\
\indent In recent years, V2X cooperative prediction has been widely applied to alleviate incomplete data problems\cite{refe11*},\cite{refe13*} caused by occlusion\cite{refe3*},\cite{refe4*} and limited lines of sight \cite{refe1*},\cite{refe2*},\cite{refe4*}. By establishing multi-source information collaboration networks among ego vehicles, infrastructure, and other traffic participants, cooperative prediction significantly improves trajectory prediction performance. However, the improvement of prediction accuracy relies on the comprehensive acquisition of multi-source data as well as a high level of accuracy and consistency. In multi-object tracking tasks, the ID switch problem--where the same traffic participant is erroneously assigned different identities in different time frames, or distinct participants are mistakenly regarded as the same identity--undermines the consistency of cross-view matching, and, when amplified during cross-view information fusion, substantially degrades the precision and robustness of cooperative trajectory prediction. Therefore, developing cooperative prediction mechanisms that ensure identity consistency and data quality has become a pivotal research direction for advancing high-accuracy trajectory prediction.\\
\indent In Vehicle-to-Everything (V2X) trajectory prediction, vehicles and roadside sensing devices establish a global situational awareness through multi-source data fusion. However, a large amount of information inevitably introduces redundant interactions and additional computational overhead, which may impair prediction performance and reduce inference efficiency. Therefore, how to fully exploit multi-source information from ego vehicle, infrastructures and other vehicles--while ensuring data integrity and reducing computational cost--has become a key research direction for improving the accuracy and efficiency of V2X trajectory prediction. Most existing V2X models \cite{refe11*},\cite{refe13*},\cite{refe10} mainly rely on maps and historical trajectories for spatio-temporal feature extraction, while neglecting the influence of traffic signals. To address this limitation, several studies \cite{refe2},\cite{refe3},\cite{refe4} have introduced the state of intersection traffic signals into their algorithms, yielding promising results. Nevertheless, the majority of these methods still rely on distance-based interaction adjustment mechanisms, failing to fully explore the potential of traffic signals in constraining effective interaction ranges and suppressing irrelevant information propagation. This oversight leads to pervasive redundant cross-agent interactions, which reduce interaction quality and significantly increase computational burden. More critically, mainstream methods \cite{refe11*}, \cite{refe13*}, \cite{refe10} often adopt vehicle-centric encoding strategies, leading to repeated encoding of shared information and overlapping historical trajectories \cite{refe16*}, \cite{refe17*}, \cite{refe19*}. The problem initially results in computational resource wastage and diminished encoding efficiency, which in turn degrades inference performance; In large-scale traffic scenarios, this degradation markedly intensifies the challenges associated with real-time deployment of the model.

The discussion highlights several challenges that impact the overall performance of V2X trajectory prediction in existing studies \cite{refe49*},\cite{refe50*}. First, in high-density traffic, frequent agent identity switches hinder accurate matching and fusion of multi-source data. Second, redundant interaction relationships and trajectory encoding hinder the precision and efficiency of spatio-temporal modeling. To address these issues, this paper proposes V2X-RECT, a trajectory prediction framework specifically designed for high-density collaborative environments. The multi-source identity matching and correction module is introduced to enhance data quality and optimize trajectory fusion. During encoding, a strategy based on a specific spatio-temporal coordinate system is employed, where traffic signal states guide agents in efficient spatio-temporal encoding. Notably, the specific spatio-temporal coordinate system enables the efficient reuse of historical trajectory features during the encoding, significantly improving encoding efficiency. Additionally, traffic signal states assist in filtering out ineffective interaction relationships, improving the quality and efficiency of encoding. Based on the spatio-temporal features and identity matching results, multi-source trajectory information is fused. A two-stage decoding strategy is then proposed to generate multiple high-quality trajectory predictions. Extensive experimental results demonstrate that the proposed method achieves significant improvements over existing state-of-the-art approaches on multiple benchmark datasets, delivering superior performance in both prediction accuracy and efficiency.

In conclusion, our main contributions are three-fold:
\begin{itemize}
\item[$\bullet$] We propose a unified trajectory prediction framework specifically designed for high-density V2X scenarios, addressing critical bottlenecks in real-time performance and robustness. The framework fundamentally mitigates identity-switch issues arising from perception errors and integrates traffic signal semantics to suppress interference from low-correlation interactions. Furthermore, it introduces a cross-temporal feature caching and reuse mechanism, forming a trajectory prediction method that balances scalability for large-scale environments with rapid real-time responsiveness.
\item[$\bullet$] In V2X scenarios, this study introduces an identity matching and correction module, which analyzes the spatio-temporal overlap across multi-view agents to address ID switching on a frame-by-frame basis and establish cross-view identity matching. In addition, the same traffic signal can assist the spatio-temporal attention mechanism in filtering the interaction relationships among agents and modeling their behavioral patterns. Finally, by decoupling the spatial relationships between scene elements and the ego vehicle, the proposed framework enables feature caching and reuse, thereby supporting parallel decoding for multiple agents.
\item[$\bullet$] Extensive experimental results on the real-world datasets (\textit{i.e.}, V2X-Seq \cite{refe11*} \& V2X-Traj \cite{refe13*}) demonstrate that our V2X-RECT achieves state-of-the-art performance in minADE, minFDE, and MR metrics. Additionally, it maintains robust performance under various traffic density conditions with a significant improvement in inference efficiency. Ablation studies further confirm the critical role of each module within the overall framework.
\end{itemize}
\section{Related Work}
\subsection{V2X Trajectory Prediction}
Single-view trajectory prediction \cite{refe39*},\cite{refe42*},\cite{refe2} has long been constrained by issues including occlusion and limited fields of view, which significantly impair model performance. With the release of several public cooperative perception datasets (\textit{e.g.} \cite{refe11*},\cite{refe13*},\cite{refe21*}), cooperative trajectory prediction has emerged as a promising and increasingly popular research direction for addressing these challenges \cite{refe11*} constructed the first real-world V2I dataset for trajectory prediction research. They manually addressed incomplete vehicle-side trajectories caused by occlusion and limited line of sight through trajectory association and stitching techniques. Although this approach is intuitive, it fails to fully exploit the motion behaviors captured from each individual view, leading to suboptimal performance. To overcome this limitation, \cite{refe13*} encodes the trajectory information from each view as independent graph nodes and introduce a cross-attention module to integrate embeddings of associated nodes from different perspectives, thereby minimizing potential information loss. While their model supports end-to-end training, the node association process requires pretraining, resulting in the formulation of two separate optimization objectives. Building on this work, \cite{refe6} proposes a cross-graph attention mechanism to fuse multi-view information without additional training. Although existing studies have made significant progress in addressing occlusion and field-of-view limitations, they often overlook issues(\textit{e.g.}, ID Switch) generated by the tracking module, which severely disrupt the accurate matching and fusion of multi-source data. We first utilize an identity matching algorithm based on multi-source trajectory information to correct ID Switch and establish the identity mapping relationships of agents.

\subsection{Traffic Signals for Trajectory Prediction}
It is well established that traffic signal states exert a profound influence on driving behavior at intersections \cite{refe2},\cite{refe3},\cite{refe23*}. While earlier work like \cite{refe23*} focused limitedly on vehicle speed variations, modern approaches necessitate a holistic view of trajectory changes. To address this issue, some studies have utilized one-hot encoding to predict vehicle intentions \cite{refe4} and simulate vehicle behaviors \cite{refe3}. Zhang et al. \cite{refe1} addressed the discontinuities in vehicle behavior caused by traffic lights by combining a spatial dynamic interaction graph (SDG) with a behavior dependency graph (BDG) in their D2-TPred model. However, the aforementioned method only focuses on the impact of phase-based signals on the future trajectory changes of agents, without delving into how continuous signals influence trajectory changes. As a result, \cite{refe1} explores the impact of continuous signals on the motion strategies of agents and proposes a mechanism to encode continuous signals while adaptively handling heterogeneous traffic control configurations. In summary, although existing research has made impressive progress in modeling the impact of traffic signals on vehicle behavior, they have overlooked the role of traffic lights in filtering interactions between agents. To bridge this gap, this paper leverages signal encoding not just as an input feature, but as a guiding signal for the attention mechanisms. By integrating signal features into both historical and agent interaction attention mechanisms, our method effectively filters out invalid interactions defined by signal phases, leading to more accurate and socially consistent trajectory predictions.

\subsection{Scene Context Encoding}
Scene context encoding enhances environmental understanding by integrating multi-dimensional information. Early single-view approaches encoded scene context into multi-channel images and employed convolutional neural networks (CNNs) to extract features. However, they suffered from inherent limitations such as restricted receptive fields, information loss, and high computational complexity. Subsequent studies have addressed these issues by adopting scene vectorization techniques, which significantly reduce computational costs while improving traffic scene feature aggregation \cite{refe15*},\cite{refe24*},\cite{refe25*}. Recently, factorized attention mechanisms\cite{refe26*},\cite{refe27*},\cite{refe28*}, exemplified by Hierarchical Vector Transformer (HIVT) \cite{refe29*}, have demonstrated remarkable performance improvements. As a result, most V2X trajectory prediction models \cite{refe11*},\cite{refe13*},\cite{refe6} have been developed based on HIVT. However, the computational cost of factorized attention mechanisms presents significant challenges for autonomous driving systems. Notably, in the field of single-view trajectory prediction, Query-Centric Network (QCNet) \cite{refe16*} introduces a query-centric scene encoding paradigm \cite{refe17*},\cite{refe18*}. This paradigm not only allows for the reuse of historical computations but also enables parallel decoding of multi-agent trajectories by sharing invariant scene features, greatly alleviating the computational burden in prediction tasks. In this study, we extend the query-centric encoding paradigm and invariant scene feature sharing method to the task of multi-view trajectory encoding.
\begin{figure*}[!t]
\centering
\includegraphics[width=6.8in]{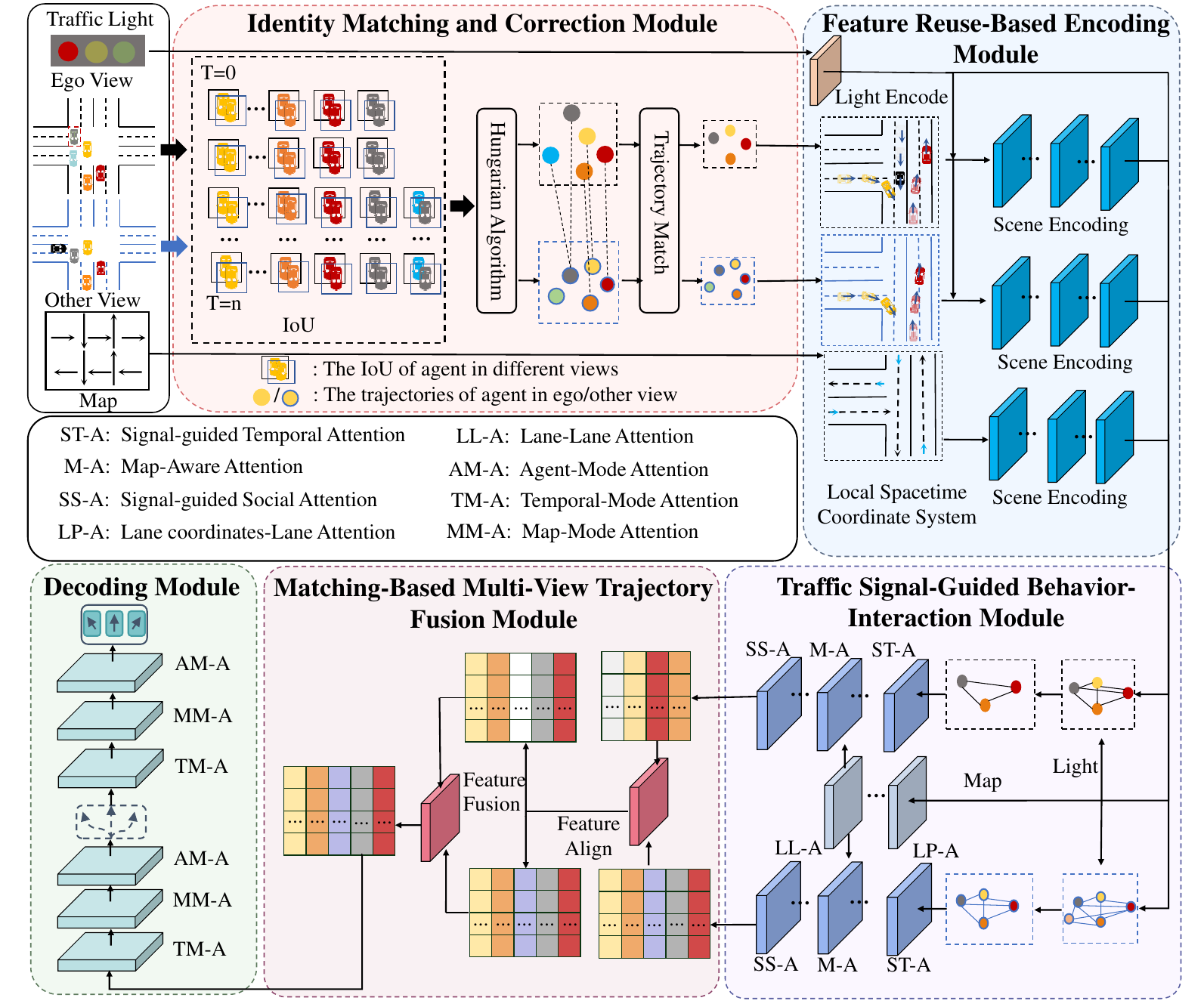}%
\hfil
\caption{The overall framework of V2X-RECT is a multi-view trajectory prediction model designed for cooperative driving environments, particularly suited to high-density traffic scenarios. The framework integrates five key mechanisms: (a) an identity matching and correction module to enhance data quality and optimize the V2X trajectory fusion; (b) a feature reuse-based encoding module to improve encoding efficiency and enable parallel computation; (c) a traffic signal-guided behavior-interaction module for modeling the impact of traffic signals on agent behavior and interactions; (d) a matching-based multi-view trajectory fusion module, which improve cross-view trajectory consistency and accuracy; (e) a decoding module that utilizes a recurrent, anchor-free proposal generator to produce adaptive trajectory anchors, and refine these initial proposals based on anchors.}
\label{fig1}
\end{figure*}
\section{Methodology}
\subsection{Problem Formulation}
The cooperative trajectory prediction module takes the historical state information $T$ of multiple agents over $t$ time steps and the map information $M$ as input $S=\{T,M\}$, and predicts $K$ trajectories \(\small T{'_i} = \{ [T{'_{{i_1}}}, \ldots ,T{'_{{i_k}}}], \ldots ,[T{'_{{{(i + n - 1)}_1}}}, \\ 
\ldots ,T{'_{{{(i + n - 1)}_k}}}]\}\) for each target agent over the next $t'$ time steps. The definitions of $T$ and $M$ are as follows:\\
\indent (1) Agent state information $T$. The multi-view agent state information can be represented as $T = \{ {T_\text{Ego}},{T_\text{Other}}\}$, where ${T_\text{Ego}}$ denotes the agent state information captured by the ego vehicle, and ${T_\text{Other}}$ denotes the state information provided by other cooperative devices (such as infrastructure and other vehicles). For agent $i$ at time step $t$, the state information includes: spatial position ${P^{t,i}} = (p_x^{t,i},p_y^{t,i})$, angular position ${\theta ^{t,i}}$ (i.e., yaw angle), velocity ${V^{t,i}} = (v_x^{t,i},v_y^{t,i})$, their bounding boxes (length, width, height), and types (pedestrian, bicycle, vehicle). The total number of states is ${N_t} = {N_\text{Ego}} + {N_\text{Other}}$. Due to occlusion and other factors, some observations may be missing in certain frames.\\
\indent (2) Map information $M$. The map information includes both map and traffic light information. The map is represented by $M$ polygonal regions in the high-definition map (such as lanes and crosswalks), with each map polygon annotated with sampling points and semantic attributes (\textit{e.g.}, lane user type). The traffic light information is recorded at a frequency of 10 Hz and includes the timestamp, location, orientation, corresponding lane ID, color state, and remaining time.
\subsection{Overall Framework}
As shown in Fig.\ref{fig1}, we propose V2X-RECT, a trajectory prediction framework, specifically designed for high-density cooperative scenarios. The framework enhances data quality by integrating multi-source trajectory information and further improves prediction accuracy and efficiency by eliminating redundant interactions and repetitive trajectory encoding. First, an identity matching and correction module is designed to address the issue of ID switching and establish identity matching relationships. Building upon this, we introduce a feature reuse-based encoding module, which avoids redundant encoding of historical trajectories and environmental information, thus improving algorithm efficiency. Furthermore, a signal-guided behavior-interaction module is proposed to model the spatio-temporal relationships of agents with the assistance of traffic signals, improving both the efficiency and accuracy of the algorithm. Subsequently, a matching-based multi-view trajectory fusion module is utilized to enhance cross-view trajectory consistency and accuracy. Finally, a decoding module is designed to decode the future trajectories of the agents.
\subsection{Identity Matching and Correction Module}
\begin{algorithm}[t!]
\caption{Trajectory Matching and Correction}
\label{alg:trajectory_matching}
\begin{algorithmic}[1]
\State \textbf{Input:} Ego-view and other-view trajectories $T_{\text{Ego}}, T_{\text{other}}$
\State \textbf{Output:} identity matching results $Map_{t_c}, Map_{t_r}$
\For{each time $t$ in $[0, T]$}
    \State Calculate Binary IoU Matrix $Binary\_Matrix \leftarrow {\mathop{\rm int}} ({M_t} \in \mathbb{R}^{{(|{T_{\text{Ego}}^t}| \times |T_{\text{other}}^t|)}} > {\tau _{{\rm{IoU}}}})$ between ego and other-view agents $T_{\text{Ego}}^t, T_{\text{other}}^t$ 
    \State $Matched\_Pairs[t] \gets$ Solving the optimal matching: Hungarian($Binary\_Matrix$)
\EndFor
\For {each $(x,\{y_1, y_2, \dots, y_n\})$ in Matched\_Pairs}
    \If {${y_{i,\text{overlapFrames}}}/{y_{i,\text{totalFrames}}} \ge \tau_{\text{overlap}}$}:
        \If {$y_i \in t_e$}
            $Map_{t_e}[t_o] \gets y_i$
        \Else:
            $Map_{t_o}[t_e] \gets y_i$
        \EndIf
    \EndIf
\EndFor
\For{each $\{y_1, y_2, \dots, y_n\}$ where $n > 1$:}
    \If $\{y_1, y_2, \dots, y_n\}$ are spatially adjacent:
        \State split $x$ into multiple trajectories $x^{new1}, \dots, x^{newn}$
    \Else:
        \State merge all $\{y_1, y_2, \dots, y_n\}$ into new trajectory $y^{new}$
    \EndIf
    \State update $Map_{t_o}$, $Map_{t_e}$
\EndFor
\end{algorithmic}
\end{algorithm}
To address the prevalent issues in object identity recognition and matching errors in existing methods, we propose an identity matching and correction module for V2X scenarios, as outlined in Algorithm~\ref{alg:trajectory_matching}. This module analyzes the spatio-temporal overlap of agents from different views, aiming to achieve two key objectives: (1) correcting ID switches to improve the quality of raw data, and (2) generating identity matching relationships to support trajectory fusion.\\
\indent Specifically, as shown in Algorithm~\ref{alg:trajectory_matching}, to analyze the spatio-temporal overlap of agents from different viewpoints, we calculate the Intersection over Union (IoU) of trajectories between the ego-vehicle and other-views at each timestamp $t \in [0, T]$ (lines 4 of Algorithm~\ref{alg:trajectory_matching}). Due to the positioning errors of trajectories from different views, one-to-many matching cases may occur. To address this, we apply the Hungarian algorithm at each timestamp $t$ to achieve optimal one-to-one matching within each time frame based on the Intersection over Union (IoU) (lines 5 of Algorithm~\ref{alg:trajectory_matching}). \\
\indent However, multi-view trajectory errors and ID switch can lead to one-to-many matching at the trajectory level, resulting in a relation $(x, \{y_1, y_2, \dots, y_n\})$, where $x \in T_{\text{ego}}$, $\{y_1, y_2, \dots, y_n\} \subseteq T_{\text{Other}}$, and $n > 1$.\\
\indent To mitigate the effect of trajectory errors, as shown in Algorithm~\ref{alg:trajectory_matching} (lines 7-13), we retain matching relation if the overlap rate exceeds a predefined threshold $\tau_{\text{overlap}}$, where the overlap rate is defined in Algorithm~\ref{alg:trajectory_matching} (lines 8).\\
\indent To correct the ID switch issue, as shown in Algorithm~\ref{alg:trajectory_matching} (lines 14-21), we first examine the coexistence of candidate targets $\{y_1, y_2, \dots, y_n\}$ within the same time frame. If multiple targets in the set are detected to occur simultaneously, we infer that the upstream task may have incorrectly merged multiple traffic participants into one object. Conversely, if the traffic participants in the set show no temporal overlap, we infer that the trajectory of a single traffic participant may have been mistakenly split into multiple independent targets. Based on this assessment, we perform trajectory splitting or merging and update the identity matching relationships.
\subsection{Feature Reuse-Based Encoding Module}
\subsubsection{Local Spacetime Coordinate System}
Typical encoding strategies first apply a temporal network to compress the time dimension, then perform agent-map and agent-agent fusion at the current time step, and subsequently fuse the agent features from the ego vehicle and other views. However, this method may result in the loss of significant information during encoding and multi-view information fusion. such as how the relationships between agents and map elements evolve over the observation horizon, or the key contextual information used to supplement occluded data during the fusion stage, which reduces both encoding and fusion quality.\\
\indent In trajectory prediction tasks, the first step is typically to encode the scene input. Inspired by recent factorized attention methods \cite{refe27*}, \cite{refe28*}, \cite{refe29*}, we observe that this approach can efficiently capture multi-dimensional dynamic features by modeling temporal relationships, agent-map interactions, and social interactions among agents along different axes separately. Building on this advantage, we further propose to perform fusion at each historical time step to more finely characterize the evolution of agent-map element relationships throughout the entire observation period, which is particularly important for V2X (Vehicle-to-Everything) trajectory encoding and multi-view trajectory fusion. However, since the computational complexity of each fusion operation is cubic, this method faces significant computational overhead and scalability limitations in high-density V2X scenarios.\\
\indent Based on the query-centered encoding paradigm in the single-agent domain \cite{refe16*}, we extend this approach to the V2X scenario to reuse redundant information. Specifically, we establish a local spatio-temporal coordinate system for each scene element and decouple it from the ego vehicle's current position. For the agents in the scene, the state of the $i$-th agent at time step $t$ is determined by the spatio-temporal position $\left(\boldsymbol{p}^{t, i}, t\right)$ and direction $\boldsymbol\theta^{t, i}$, where $\boldsymbol{p}^{t, i}$ is the spatial position of the agent. For lanes and crosswalks, the position and direction at the centerline entrance are chosen as references. We use a four-dimensional descriptor to summarize the relative position of agent $i$ at time $t$ and agent $j$ at time $s$, whose components include relative distance $\left|\boldsymbol{p}^{s, j}-\boldsymbol{p}^{t, i}\right|$, relative direction $\arctan \left(\boldsymbol{p}_{y}^{s, j}-\boldsymbol{p}_{y}^{t,i}, \boldsymbol{p}_{x}^{s, j}-\boldsymbol{p}_{x}^{t, i}\right)-\boldsymbol{\theta}^{t, i}$, relative orientation $\boldsymbol{\theta}^{s, j}-\boldsymbol{\theta}^{t, i}$, and time difference $s - t$.

\subsubsection{Transition Trend of Traffic Signals}
In a traffic environment, the changing trend of traffic signals has a greater impact on the behavior of agents than the current signal state. For example, when a green light is about to end, vehicles at an intersection may choose to accelerate through the intersection or slow down to stop in advance. Therefore, we extract the changing trend of traffic signals to assist in prediction. Specifically, in this module, we first identify the road areas controlled by traffic signals and assign a value of -999 to the signal information of agents not located within these areas. For agents on the controlled road areas, we extract the changing trend of the traffic signal based on the continuous numerical signal data they receive. The changing trend of the traffic signal can be formulated as:
\begin{equation}
P{E_{t,S}} = \arctan \left( {{{{t_{d,{S_{{\rm{reamin }}}}}}}}/{{{T_\Omega }}} \cdot \frac{1}{3}d} \right),
\end{equation}
where $t_{d, S_{\text {remain}}}$ represents the remaining time of each color signal of a single traffic light $S$, $T_{\Omega}$ refers to the maximum cycle time of the traffic light sequence, and $d\in \{0, 1, 2\}$ indicates the color of signals, following the specific order of red, green, and yellow.

\subsubsection{Encoding of Scene Elements}
\indent For traffic participants observed from different views, the raw trajectory data has a low dimensionality and contains limited information. To address this, we first apply Fourier transform \cite{refe30*},\cite{refe31*},\cite{refe32*} to map the raw data from the time domain to the frequency domain, extracting richer frequency-domain features. Then, a multi-layer perceptron (MLP) is used to map the frequency-domain data to a higher-dimensional space, enabling the extraction of abundant information suitable for the subsequent prediction can be extracted. Specifically, we obtain the state information of the traffic participants $a^{i,m} = \left\{ {p_{m}^{t,i},\theta _{m}^{t,i},v_{m}^{t,i},c_{m}^i,P{E_{t,S}},t} \right\}$ at time $t$, Then, for different data sources, we apply Fourier transform extract trajectory of the agents features in the frequency domain, which can be formulated as:
\begin{equation}
f_{a,m}^t = {F_{{\rm{Fourier}}}}\left( {p_m^t,\theta _m^t,v_m^t,P{E_{t,S}},t} \right),
\end{equation}
where $m=0$ indicates the data from the ego vehicle and $m=1$ indicates the data from other views, $v$ indicates velocity of agent $a$. Therefore, the frequency-domain feature sequence can be expressed as $f_{a,m} = [f_{a,m}^1,f_{a,m}^2, \ldots ,f_{a,m}^T]$.\\
\indent Since the scene information remains unchanged across different views, we encode the lane information $l = \left\{ {p,c,t} \right\}$ and then share the feature across all vies. The Fourier transform of the lane information can be formally represented as:
\begin{equation}
{f_{l,j}} = {F_{{\rm{Fourier}}}}\left( {{p_j},t} \right),
\end{equation}
where $p_j$ indicates the position of lane $j$. Finally, for each agent state and each sampled point on the map, the corresponding Fourier features are concatenated with their semantic attributes (e.g., category). These combined representations are then processed by MLP to generate their embeddings $f_{a,i}$, $f_{l,j}$ respectively.  
\indent For two agents controlled by the same traffic signal, with the spatial-temporal position of agent $i$ $(p_i^t,\theta_i^t,t)$ and agent $j$ having $(p_j^s, \theta_j^s, s)$, (if one of them is a map element, we do not consider the impact of the traffic signal on both.) We use a 4D descriptor to summarize their relative position. Then, we transform the 4D descriptor into Fourier features and pass them through an MLP to produce the relative positional embedding $r_{j \to i}^{{\rm{s}} \to {\rm{t}}}$. The relative position embeddings for scene element pairs can be formulated as:
\begin{equation}
\begin{aligned}
    f_{j \to i}^{s \leftarrow t} = {F_{{\rm{Fourier}}}}(\left| {{p^{s,j}} - {p^{t,i}}} \right|,{\theta ^{s,j}} - {\theta ^{t,i}},\;s - t, \\ \arctan \left( {p_y^{s,j} - p_y^{t,i},\;p_x^{s,j} - p_x^{t,i}} \right) - {\theta ^{t,i}}),
\end{aligned}
\end{equation}
where $F_{{\rm{Fourier}}}$ indicates the Fourier transformer module, If any of the two scene elements are static (\textit{e.g.}, static map polygons), we can omit the superscript and denote the embedding as $r_{j \rightarrow i}$.

\subsection{Signal-Guided Behavior-Interaction Module}
In this section, we propose a signal-guided behavioral-interaction module, which models the impact of traffic signals on agent behavior and interactions through factorized attention \cite{refe27*},\cite{refe28*},\cite{refe29*}. As shown in Fig.\ref{fig1}, the mechanism consists of five core modules: Lane and Lane Position Attention (LP-A): Designed to capture the mapping between lane coordinate points and their associated lanes; Lane-to-Lane Attention (LL-A): Designed to capture the topological dependencies among different lanes; Signal-guided Temporal Attention (ST-A), which models the impact of traffic signals on agent behavior; Map-aware Attention (M-A), which extracts the positional information of agents on the map; And Signal-guided Social Attention (SS-A), which models the impact of traffic signals on agent interaction relationships.  
\subsubsection{Lane and Lane Position Attention (LP-A)}
We utilize a self-attention mechanism to capture the relationships between lane coordinate points and their associated lanes, which can be formulated as:
\begin{equation}
{f_{l,i}} = {\mathop{F_{{\rm{LP-A}}}}\nolimits} \left( {{[f_{pt,i},f_{pl,i}]},\left[ {{f_{pl,i}},{r_{pt,i \to pl,i}}} \right]} \right),
\end{equation}
\begin{equation}
{Q_{pl,i}} = W_Q^{Map}[f_{pt,i},f_{pl,j}],
\end{equation}
\begin{equation}
{K_{pl,i}} = W_K^{Map}\left[{f_{pl,i},{r_{pt,i \to pl,i}}}\right],
\end{equation}
\begin{equation}
{V_{pl,i}} = W_V^{Map}\left[{f_{pl,i},{r_{pt,i \to pl,i}}}\right],
\end{equation}
where $f_{pt,i}$ denotes the position embedding of the $i$-th lane, $f_{pl,i}$ denotes the embedding of the $i$-th lane polygon, which are used as the query vector $Q_{pl,i}$. The embeddings $f_{pl,i}$ of $i$-th lane polygon is combined with relative positional encodings $r_{pt,i \to pl,i}$ to construct the key $K_{pl,i}$ and value $V_{pl,i}$.
\subsubsection{Lane and Lane Attention (LL-A)}
To model the relationships between map elements, similar to \cite{refe16*}, which can be formulated as:
\begin{equation}
{x_{pl,i}} = {\mathop{F_{{\rm{M-A}}}}\nolimits} \left( {{f_{l,i}},\left[ {{f_{l,j}},{r_{j \to i}}} \right]} \right),
\end{equation}
\begin{equation}
{Q_{pl,i}} = W_Q^{Map}f_{l,i},
\end{equation}
\begin{equation}
{K_{pl,i}} = W_K^{Map}\left[{f_{l,j},{r_{j \to i}}}\right],
\end{equation}
\begin{equation}
{V_{pl,i}} = W_V^{Map}\left[{f_{l,j},{r_{j \to i}}}\right],
\end{equation}
where $f_{l,i}$ is used as the query vector $Q_{pl,i}$. The embeddings $f_{l,j}$ of adjacent polygons are combined with relative positional encodings $r_{j \to i}$ to construct the key $K_{pl,i}$ and value $V_{pl,i}$ , thereby allowing the encoder to explicitly incorporate spatial relationships between map elements. Since each input to the attention layer is independent of the global spacetime coordinate system, the output map encodings are also invariant under transformations of the global reference frame. Thus, they can be shared across all agents and all time steps and can even be precomputed offline, thereby avoiding redundant computation suffered by agent-centric modeling.
\subsubsection{Signal-Guided Temporal Attention (ST-A)}
To effectively model the temporal evolution of motion patterns influenced by the traffic signal change trends, we concatenate the state embedding of agent $i$ at the current moment with the traffic signal change trends and input this into a MLP to obtain the agent's state feature. This agent state feature is then used to generate the query vector. The key and value vectors are derived by combining the agent's state over the historical time window $[t-\tau,t-1]$, along with its temporal relative position encoding. The process is formulated as follows:
\begin{equation}
{f_{a,i}} = {\mathop{F_{{\rm{ST-A}}}}\nolimits} \left( {f_{a,i}^t,\left[ {f_{a,i}^{t - T},{r^{t - T \to t}}} \right]} \right),
\end{equation}
\begin{equation}
{Q_{a,i}} = W_Q^{Temporal}f_{a,i}^t,
\end{equation}
\begin{equation}
{K_{a,i}} = W_K^{Temporal}\left[{f_{a,i}^{t - T},{r^{t - T \to t}}}\right],
\end{equation}
\begin{equation}
{V_{a,i}} = W_V^{Temporal}\left[{f_{a,i}^{t - T},{r^{t - T \to t}}}\right],
\end{equation}
where $f_{a,i}^t$ denotes the current state embedding of agent $i$, $f_{a,i}^{t - T}$ represents its historical states, and $r^{t - T \to t}$ indicates the temporal relative position encoding. $Q_{a,i}$, $K_{a,i}$ and $V_{a,i}$ denote query, key and value vectors, respectively.

\subsubsection{Map-Aware Attention(M-A)}
To assist the agent in obtaining positional relationships in the map, we generate a query vector based on the state feature of agent $i$ at the current moment, and then generate a key/value vector by fusing geometry features of the lane and relative position embedding, therefore establishing an explicit representation of the agent-lane topological association. In this way, map-aware attention can be formulated as follows: 
\begin{equation}
f_{a,i} = {\mathop{F_{{\rm{M-A}}}}\nolimits} \left( {f_{a,i}^t,\left[ {x_{pl,j}^t,{r_{j \to i}}} \right]} \right),
\end{equation}
where ${f}_{a,i}^{t}$ denotes the state feature of agent $i$ at the current time step, $x_{pl,j}^{t}$ represents the geometric features of lane $j$, and $r_{j \to i}$ represents the relative position embedding from lane $j$ to agent $i$. The computation process for the query, key and value vectors are identical to that described earlier, differing only in the input or parameter settings.

\subsubsection{Signal-Guided Social Attention (SS-A)}
To effectively model the interactions between agents influenced by signal states, for agents $i$ and $j$ controlled by the same traffic signal, we generate a query vector using the position encoding of agent $i$, and inject the position encoding of agent $j$ along with the signal-guided spatial neighborhood relationship into the key/value vector generation process. Signal-guided social attention can be formulated as follows:
\begin{equation}
{f_{a,i}} = {\mathop{F_{{\rm{SS-A}}}}\nolimits} \left( {f_{a,i},\left[ {f_{a,j},{r_{j \to i}}} \right]} \right),
\end{equation}
where $f_{a,i}$, and $f_{a,j}$ denote the position encodings of agents $i$ and $j$, respectively, and $r_{j \rightarrow i}$ represents the signal-guided spatial relation from agent $j$ to agent $i$. The computation process of query, key and value are identical to that described earlier, differing only in the input or parameter settings.

\subsection{Matching-Based Multi-View Trajectory Fusion}
\subsubsection{Spatio-Temporal Feature Alignment and Fusion}
Due to certain errors in the information between the vehicle and other views, directly supplementing the trajectory information may lead to error accumulation, thus preventing the effective utilization of the complete trajectory information and affecting the performance of downstream tasks. To solve the issue of spatio-temporal inconsistency in the multi-view information fusion, this paper proposes a feature enhancement method based on spatio-temporal alignment.

First, trajectory feature alignment is performed. Subsequently, the spatial positions of the trajectory features are adjusted so that matched trajectories from different views are aligned to the same feature column before the attention operation, facilitating more effective cross-attention fusion across views.

Afterwards, discrete Fourier transform (DFT) is applied to trajectory feature data from different views to obtain their frequency domain representations, \textit{i.e.},
\begin{equation}
f_{a,0} = \mathcal{F_{\text{Fourier}}}({f_{a,0}}),
\end{equation}
\begin{equation}
f_{a,1} = \mathcal{F_{\text {Fourier}}}(f_{a,1}),
\end{equation}
where \( \mathcal{F_{\text {Fourier}}}(\cdot) \) denotes the Fourier transform operation on the sequences of trajectory features along the time dimension, and \( f_{a,0} \) and \( f_{a,1} \) respectively denote the trajectory features as observed from the ego-vehicle view and from other views. \\
\indent To achieve cross-modal feature fusion, a map-aware cross-attention mechanism is introduced. This mechanism uses ego vehicle-side trajectory features as query vectors, and integrates other-side trajectory features along with their relative positional embeddings as key/value vectors. To be specific, the cross-attention operation can be formulated as:
\begin{equation}
{f_{e,o}} = {\mathop{F_{{\rm{fusion}}}}\nolimits} \left( {{f_{a,0}},\left[ {{f_{a,1}},{r^{e \to o}}} \right]} \right),
\end{equation}
where ${r}_{j \rightarrow i}^{e \rightarrow o}$ indicates their relative positional embeddings. The computation process of query, key and value embeddings are identical to that described earlier, only differing in the input or parameter settings.

\subsection{Decoding}
To address common issues in single-stage decoding methods such as training instability, mode collapse, and long-term prediction bias \cite{refe26*},\cite{refe33*}, we use a query-based dual-stage decoder for v2x trajectory prediction. Specifically, an anchor-free proposal module is first introduced to generate adaptive trajectory anchors by fusing vehicle-road trajectory features and scene encodings. These initial anchors are then refined by an anchor-based optimization module to enhance the precision of the predicted trajectories. This two-stage design improves long-term prediction accuracy and robustness under complex conditions. The core processing flow of the decoding pipeline is illustrated in Fig.~\ref{fig1}, and each module is detailed below.

\subsubsection{Proposal Module}
Our proposal module generates $K$ adaptive anchors in a data-driven manner. Thanks to the cross-attention layer, the mode queries first leverage the agent's historical attention to capture the agent's motion style and analyze its potential goals. Then, thanks to the cross attention layers, the mode queries can retrieve the scene context and quickly narrow the search space for anchors. The self-attention layer further allows the queries to collaborate with each other when generating trajectory proposals. Finally, an initial trajectory proposal is generated through a MLP, serving as the trajectory anchor for subsequent refinement. It can be specifically formulated as:
\begin{equation}
{f_{e,o,i}} = {\mathop{F_{{\rm{AM-A}}}}\nolimits} \left( {f_{e,o,i}^t,\left[ {f_{e,o,i}^{t - T},{r^{t - T \to t}}} \right]} \right),
\end{equation}
\begin{equation}
f_{e,o,i}^t = {\mathop{F_{{\rm{MM-A}}}}\nolimits} \left( {f_{e,o,i}^t,\left[ {x_{pl,j}^t,{r_{j \to i}}} \right]} \right),
\end{equation}
\begin{equation}
{f_{e,o,i}} = {\mathop{F_{{\rm{TM-A}}}}\nolimits} \left( {f_{e,o,i},\left[ {f_{e,o,j},{r_{j \to i}}} \right]} \right),
\end{equation}
where the computation process of query, key and value are identical to that described earlier, differing only in the input or parameter settings.

\subsubsection{Anchor-Based Trajectory Refinement}
To further improve temporal consistency and improve trajectory prediction accuracy, a second stage motion refinement module is introduced. This stage aims to reduce the offset between predicted trajectories and ground truth, while suppressing physically implausible or behaviorally noncompliant outputs. In this stage, we adopt the output of the proposal module as an anchor, and an architecture similar to the proposal module is employed to motion refinement. These anchor-based queries provide explicit spatial priors, allowing the attention layers to better focus on relevant contextual information, thereby improving the plausibility and accuracy of the final predictions.

\subsection{Training Objective}
Similar to QCNet, we optimize the mixing coefficients predicted by the refinement module through the classification loss $L_{cls}$. Meanwhile, we adopt a "winner-takes-all" strategy \cite{refe34*} to optimize the position and scale parameters output by the proposal module and the refinement module. The final loss function combines the trajectory proposal loss $L_\text{{propose}}$, the trajectory refinement loss $L_\text{{refine}}$ and the classification loss $L_{cls}$ for end-to-end training:
\begin{equation}
L = L_{\text{propose}} + L_{\text{refine}} + \lambda \cdot L_{\text{cls}}.
\end{equation}
In this context, $\lambda$ is utilized to balance the regression and classification objectives.

\section{Experiments}
\label{sec5}

\subsection{Experimental Setup}

\subsubsection{Datasets}
We evaluate the proposed model on two large-scale, real-world V2X datasets, \textit{i.e.}, V2X-Seq \cite{refe11*} and V2X-Traj \cite{refe13*}. All datasets are divided into training and validation sets, and the trained model is evaluated on the validation set to compare with existing methods.\\
\indent V2X-Seq dataset provides agent trajectories from both the ego vehicle and infrastructure sides, as well as vector maps and traffic signal information. V2X-Seq contains 51,146 V2I scenarios, with each scenario being 10 seconds long and sampled at 10 Hz. We are tasked with predicting 5-second future trajectories based on 5 seconds of historical motion data from both the vehicle-side and infrastructure views for the V2X-Seq dataset.\\
\indent V2X-Traj dataset provides agent trajectories which provides agent trajectories from ego vehicle and other sides (other vehicle or infrastructure), vector maps, and traffic signal information. It comprises 10,102 scenarios in challenging intersections. Each scenario lasts for 8 seconds with a sample rate of 10 Hz. The 4-second observations from each view are used to predict the future motion in the next 4 seconds.

\subsubsection{Evaluation Metrics}
According to the standard evaluation protocol, we used minADE, minFDE, and MR to evaluate the model. which are can be formulated as:
\begin{equation}
    \min ADE = \min _{K = 1}^k\sum\nolimits_{t = 1}^T {{{\left\| {\hat y_{i,k}^t - y_i^t} \right\|}^2}},
\end{equation}
\begin{equation}
    \min FDE = \min _{K = 1}^k{\left\| {\hat y_i^T - y_i^T} \right\|^2},
\end{equation}
\begin{equation}
    MR = {{count({{\left\| {\hat y_i^T - y_i^T} \right\|}^2} > 2)}}/{K},
\end{equation}
where $\hat y$ denotes the predicted trajectories generated by our method, $y$ denotes the ground-truth trajectories.\\
\indent The metric minADEK computes the l2 distance (in meters) between the true trajectory and the best trajectory among K predicted trajectories as the average over all future time steps. On the other hand, the metric minFDEK focuses only on the prediction error of the last time step to emphasize long-term performance. In addition, the metric missing rate MRK measures the percentage of scenes with an error greater than 2.0 meters in the last time step. The lower the metric values, the better the prediction performance. We compute the error between the best predicted trajectory and the ground truth trajectory in K=6 modes.

\subsubsection{Implementation Details}
All the experiments are conducted on a machine with NVIDIA A100 GPUs. The overall model is trained for a total of 64 epochs using the AdamW optimizer, with an initial learning rate of $5 \times 10^{-4}$.

\subsection{Main Results}
\subsubsection{Comparison With State of the Art}
On V2X-Seq dataset, we compare the proposed V2X-RECT with the following baseline approaches.
PP-VIC \cite{refe11*} stitches the AV and infrastructure trajectories through CBMOT and then tests them using the SOAT prediction models TNT \cite{refe10*} and HiVT \cite{refe29*}. V2X-Graph \cite{refe13*} constructs a graph to fuse the historical trajectories from AV and infrastructure, however, it focuses only on the historical domain fusion. V2INet \cite{refe6} efficiently models multi-view data by extending existing single-view models. Additionally, the method incorporates a post-hoc conformal prediction module to calibrate the generated multi-modal trajectories. AIoT \cite{refe10} enables real-time information exchange and trajectory prediction of target agents by leveraging IoT sensor technologies and incorporating two layers of cooperation: ego-vehicles and infrastructures. Co-HTTP \cite{refe14*} method collaborates with autonomous vehicles by transmitting infrastructure historical data, embedding the intentions of surrounding agents into a heterogeneous graph neural network, and performing heterogeneous updates to optimize driving intention prediction and vehicle-infrastructure cooperation. Co-MTP \cite{refe15*} framework capture interactions in both history and future domains, utilizes a heterogeneous graph transformer to fuse historical features and predict future interactions, thereby optimizing planning actions and the final scenario state.
I2Xtraj \cite{refe2} leverages a continuous signal-informed mechanism to adaptively process real-time traffic signals from infrastructure devices. Based on the prior knowledge of intersection topology, a driving strategy awareness mechanism is proposed to model the joint distribution of goal intentions and maneuvers.

\begin{table}[t!]
\footnotesize
\caption{
  Cooperative method comparison on V2X-seq. * represents the published work. The best performance is in bold and the second best is underlined. C is the collaborative set, while ego is the single vehicle set.}
\label{table1}
\centering
\begin{tabular}{>{\centering\arraybackslash}m{2.8cm}>{\centering\arraybackslash}m{1.8cm}>{\centering\arraybackslash}m{0.7cm}>{\centering\arraybackslash}m{0.7cm}>{\centering\arraybackslash}m{0.7cm}}
\toprule
Method & Cooperation & minADE & minFDE & MR \\
\midrule
TNT\cite{refe10*} & ego & 8.54 & 17.93 & 0.77\\
TNT\cite{refe10*}(2024) & C/PP-VIC\cite{refe11*} & 7.38 & 15.27 & 0.72\\
HIVT\cite{refe29*} & ego & 1.34 & 2.16 & 0.31\\
HIVT\cite{refe29*}(2024*) & C/PP-VIC & 1.28 & 2.11 & 0.31\\
V2X-Graph\cite{refe13*}(2024*) & C/FF & 1.17 & 2.03 & 0.29\\
V2INet\cite{refe6}(2024) & C/FF & 1.19 & 1.98 & 0.27\\
AIoT\cite{refe10}(2024*) & C/FF & 0.95 & 1.87 & 0.27\\
Co-HTTP\cite{refe14*}(2024*) & C/FF & 0.83 & 1.33 & 0.19\\
Co-MTP\cite{refe15*}(2025.02)  & C/FF & \underline{0.76} & \underline{1.15} & 0.16\\
I2XTraj\cite{refe2}(2025.03) & C/FF & 0.77 & 1.34 & \underline{0.13}\\
V2X-RECT(Ours) & C/FF & \textbf{0.53} & \textbf{0.80} & \textbf{0.09}\\
\bottomrule
\end{tabular}
\end{table}

\indent Following the evaluation protocol in \cite{refe11*}, we benchmark our method on the V2X-Seq and V2X-Traj datasets against benchmark models, with results summarized in Table \ref{table1}. We compare our proposed method with other collaborative approaches on the V2X-Seq dataset. TNT \cite{refe10*} and HiVT \cite{refe29*} are typical single-view prediction models. In \cite{refe11*}, they are evaluated under two settings: the Ego setting, which uses only vehicle-view data, and the PP-VIC setting, a two-stage pipeline that combines vehicle-view and infrastructure-view data, while FF indicates a Feature Fusion method\cite{refe13*}. In the evaluation, besides our method, Co-MTP achieved the best performance in minADE and minFDE metrics, while I2XTraj demonstrated the best results on the MR metric. However, compared to these methods, our approach, V2X-RECT, outperformed all three evaluation metrics (minADE, minFDE, and MR). Specifically, V2X-RECT achieved a reduction of 0.23 in minADE (a 30.26\% decrease) and 0.35 in minFDE (a 30.43\% decrease) compared to Co-MTP. This can be attributed to the design of our encoding and fusion modules. Specifically, the identity matching and correction module, along with the traffic signal-guided interaction-behavior modeling module, significantly enhanced data quality and enabled more comprehensive and precise modeling of agent behaviors and underlying mechanisms. Compared to I2XTraj, our method demonstrates significant improvements across all metrics. Specifically, MR decreases by 0.04 (a 30.77\% reduction), minADE decreases by 0.24 (a 31.17\% reduction), and minFDE decreases by 0.54 (a 40.3\% reduction). Notably, for the minFDE metric, since the decoding module of Co-MTP also adopts a target-point-based decoding approach similar to our method, the advantage of our method over Co-MTP is less pronounced than the advantage over I2XTraj.

\begin{table}[t!]
\footnotesize
\caption{Method comparisons on V2X-Traj (ego vehicle).}
\label{table2}
\centering
\begin{tabular}{>{\centering\arraybackslash}m{2.2cm}>{\centering\arraybackslash}m{1.2cm}>{\centering\arraybackslash}m{1.2cm}>{\centering\arraybackslash}m{1.2cm}}
\toprule
Method & minADE & minFDE & MR \\
\midrule
DenseTNT\cite{refe10*} & 1.23 & 2.09 & 0.25\\
HDGT\cite{refe8*} & 0.91 & \underline{1.48} & \underline{0.14}\\
V2X-Graph\cite{refe13*} & \underline{0.90} & 1.56 & 0.17\\
V2X-RECT(Ours) & \textbf{0.41} & \textbf{0.52} & \textbf{0.04}\\
\bottomrule
\end{tabular}
\end{table}

\indent On the V2X-Traj dataset, besides V2X-Graph, DenseTNT, we also compare V2X-RECT with HDGT, HDGT \cite{refe8*} models heterogeneous elements and semantic relations through a heterogeneous graph, encodes relative spatial information in a local coordinate system, and adopts a hierarchical Transformer structure to adapt to heterogeneous aggregation.

\begin{table}[t!]
\footnotesize
\caption{The cooperative method comparisons on the V2X-seq (V2V) dataset.}
\label{table3}
\centering
\begin{tabular}{>{\centering\arraybackslash}m{2.2cm}>{\centering\arraybackslash}m{1.2cm}>{\centering\arraybackslash}m{1.2cm}>{\centering\arraybackslash}m{1.2cm}}
\toprule
Method & minADE & minFDE & MR \\
\midrule
DenseTNT\cite{refe10*} & 1.20 & 2.04 & 0.25\\
HDGT\cite{refe8*} & 0.94 & 1.57 & 0.17\\
V2X-Graph\cite{refe13*} & \underline{0.77} & \underline{1.26} & \underline{0.12}\\
V2X-RECT(Ours) & \textbf{0.38} & \textbf{0.41} & \textbf{0.03}\\
\bottomrule
\end{tabular}
\end{table}

\begin{table}[t!]
\footnotesize
\caption{The cooperative method comparisons on the V2X-seq (V2I) dataset.}
\label{table4}
\centering
\begin{tabular}{>{\centering\arraybackslash}m{2.2cm}>{\centering\arraybackslash}m{1.2cm}>{\centering\arraybackslash}m{1.2cm}>{\centering\arraybackslash}m{1.2cm}}
\toprule
Method & minADE & minFDE & MR \\
\midrule
DenseTNT\cite{refe10*} & 1.32 & 2.34 & 0.29\\
HDGT\cite{refe8*} & 0.94 & 1.59 & 0.16\\
V2X-Graph\cite{refe13*} & \underline{0.80} & \underline{1.30} & \underline{0.13}\\
V2X-RECT(Ours) & \textbf{0.39} & \textbf{0.43} & \textbf{0.03}\\
\bottomrule
\end{tabular}
\end{table}

\begin{table}[t!]
\footnotesize
\caption{Performance comparisons under the different traffic densities.}
\label{table5}
\centering
\begin{tabular}{>{\centering\arraybackslash}m{1.4cm}>{\centering\arraybackslash}m{2.2cm}>{\centering\arraybackslash}m{1cm}>{\centering\arraybackslash}m{1cm}>{\centering\arraybackslash}m{1cm}}
\toprule
Number of agents & Method & minADE & minFDE & MR \\
\midrule
Scene 1 & V2INet\cite{refe6} & 0.9625 & 1.9588 & 0.2899\\
$0-190$ & V2X-RECT(Ours) & 0.5303 & 0.8037 & 0.0894\\
\midrule
Scene 2 & V2INet\cite{refe6} & 1.2567 & 2.1781 & 0.3016\\
$190-290$ & V2X-RECT(Ours) & 0.5306 & 0.8043 & 0.0892\\
\midrule
Scene 3 & V2INet\cite{refe6} & 1.2633 & 2.1524 & 0.3261\\
$290-390$ & V2X-RECT(Ours) & 0.5303 & 0.8039 & 0.0901\\
\midrule
Scene 4 & V2INet\cite{refe6} & 1.5114 & 2.4847 & 0.3301\\
$390-490$ & V2X-RECT(Ours) & 0.5309 & 0.8042 & 0.0903\\
\midrule
Scene 5 & V2INet\cite{refe6} & 1.5109 & 2.4785 & 0.3382\\
$490-590$ & V2X-RECT(Ours) & 0.5304 & 0.8438 & 0.0895\\
\midrule
Scene 6 & V2INet\cite{refe6} & 1.4605 & 2.3703 & 0.3462\\
$590-$ & V2X-RECT(Ours) & 0.5305 & 0.8040 & 0.0893\\
\bottomrule
\end{tabular}
\end{table}

\indent Due to the timing of the V2X-Traj dataset release relatively late, many existing algorithms have yet to be validated on this dataset. Limited by multiple factors (such as learning rate, initial weight, and the unavailability of code for certain studies), reproducing experimental results for these methods is considerably challenging, making direct comparisons with these methods infeasible. Therefore, to evaluate the performance of the proposed method in V2X scenarios, we selected three mainstream baseline methods cited in the V2X-Graph paper for comparison experiments. As shown in Tables \ref{table2},\ref{table3},\ref{table4}, our method significantly outperforms the other methods across all cooperative settings (Ego Vehicle, V2V, and V2I). Compared to V2X-Graph, our approach demonstrates notable advantages in trajectory quality, interaction modeling, and information fusion. Additionally, by optimizing the decoding process, we further enhance prediction performance, showcasing superior performance in V2X scenarios.
\begin{figure*}[h]
    \centering
    \begin{subfigure}[b]{0.32\textwidth}
        \centering
        \includegraphics[width=\textwidth]{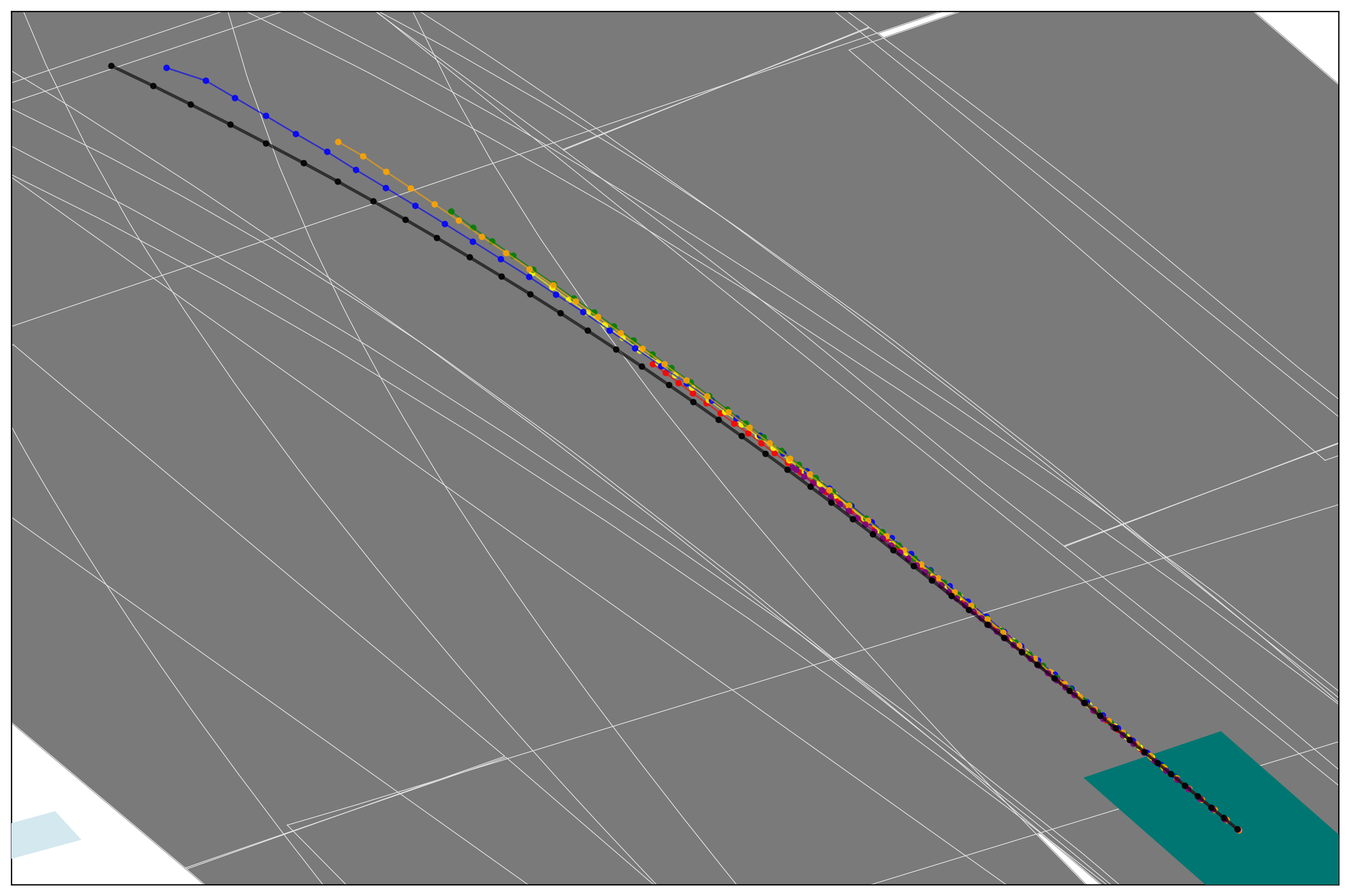}
    \end{subfigure}
    \begin{subfigure}[b]{0.32\textwidth}
        \centering
        \includegraphics[width=\textwidth]{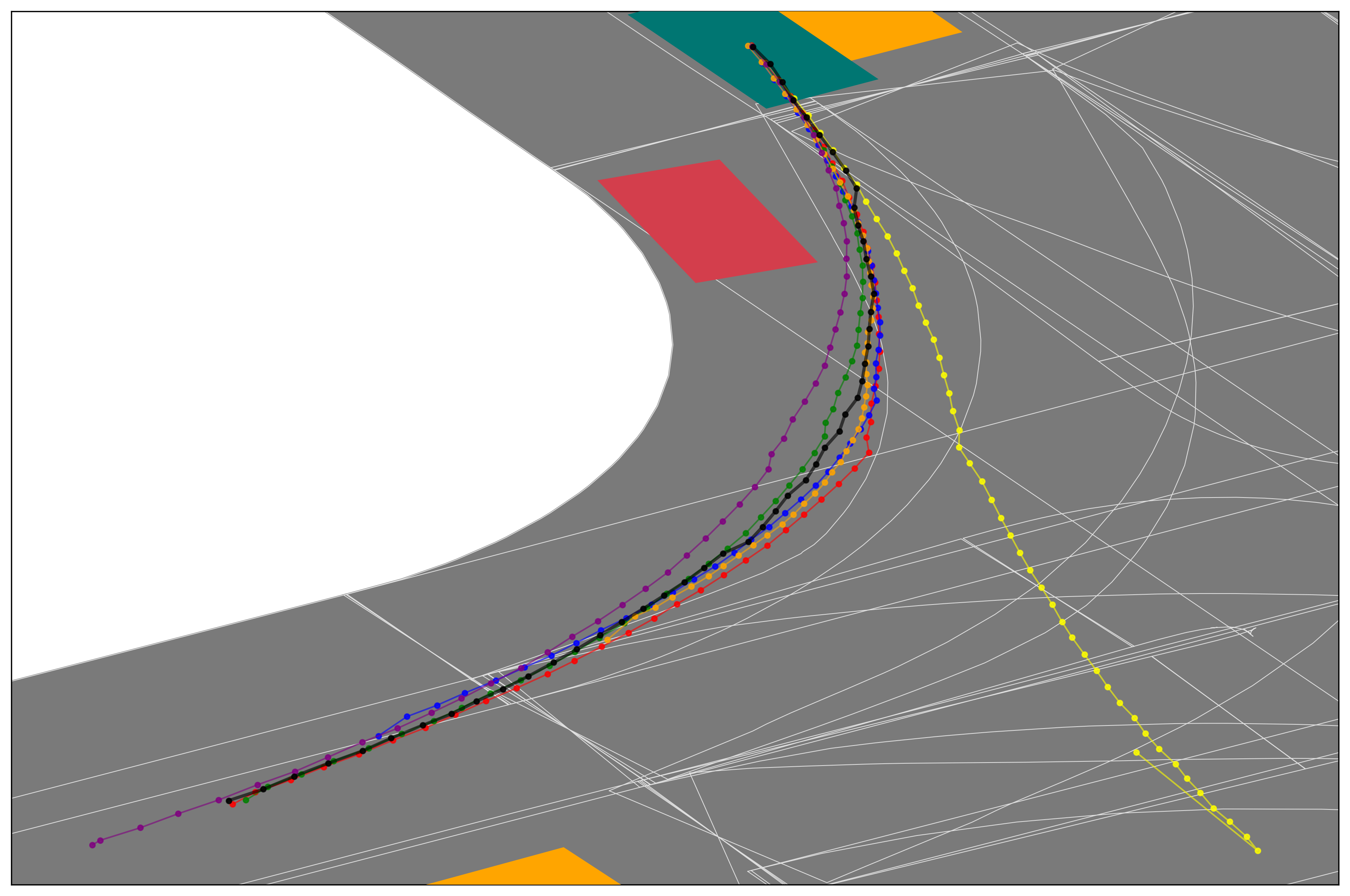}
    \end{subfigure}
    \begin{subfigure}[b]{0.32\textwidth}
        \centering
        \includegraphics[width=\textwidth]{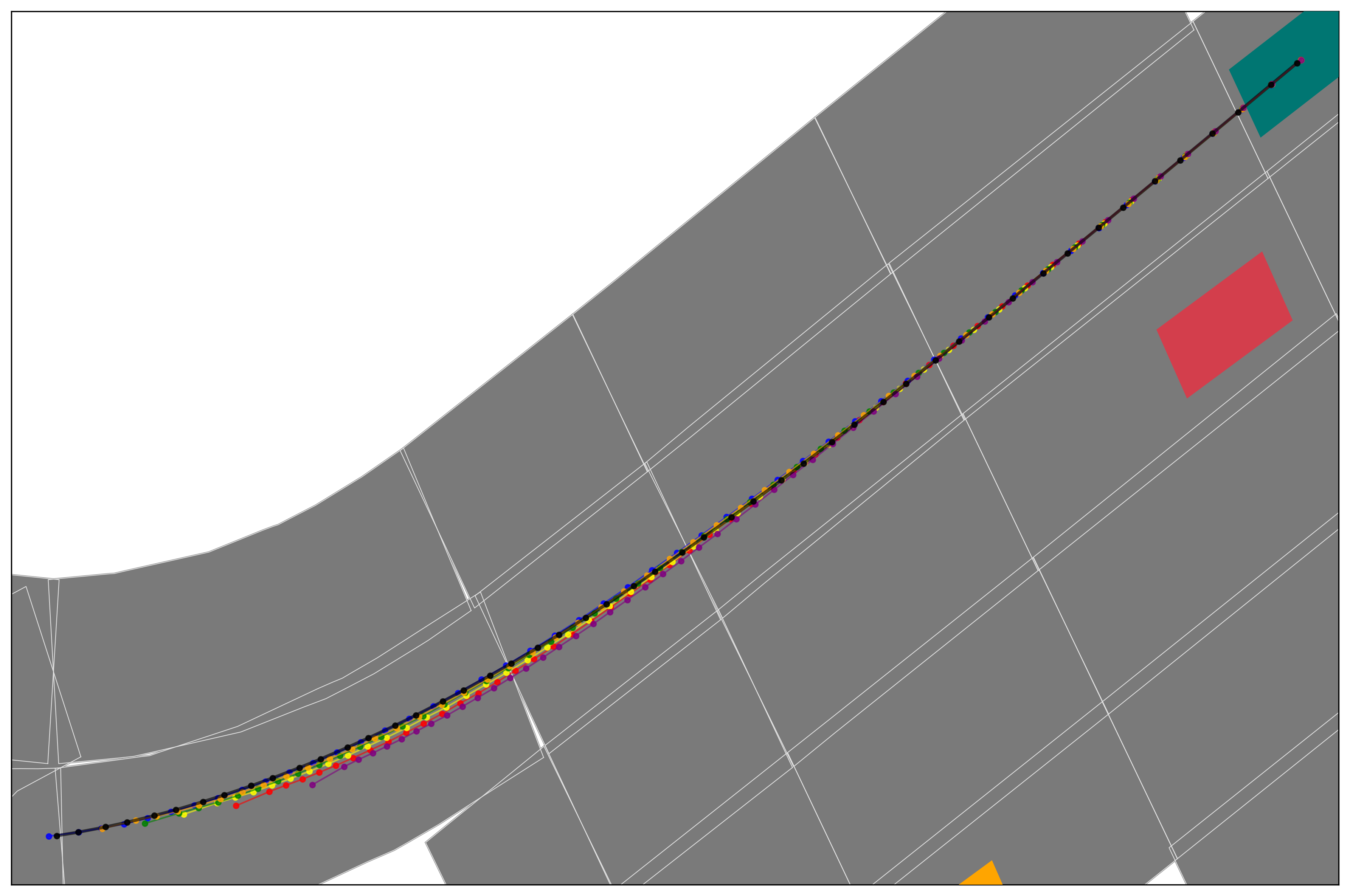}
    \end{subfigure}


    \begin{subfigure}[b]{0.32\textwidth}
        \centering
        \includegraphics[width=\textwidth]{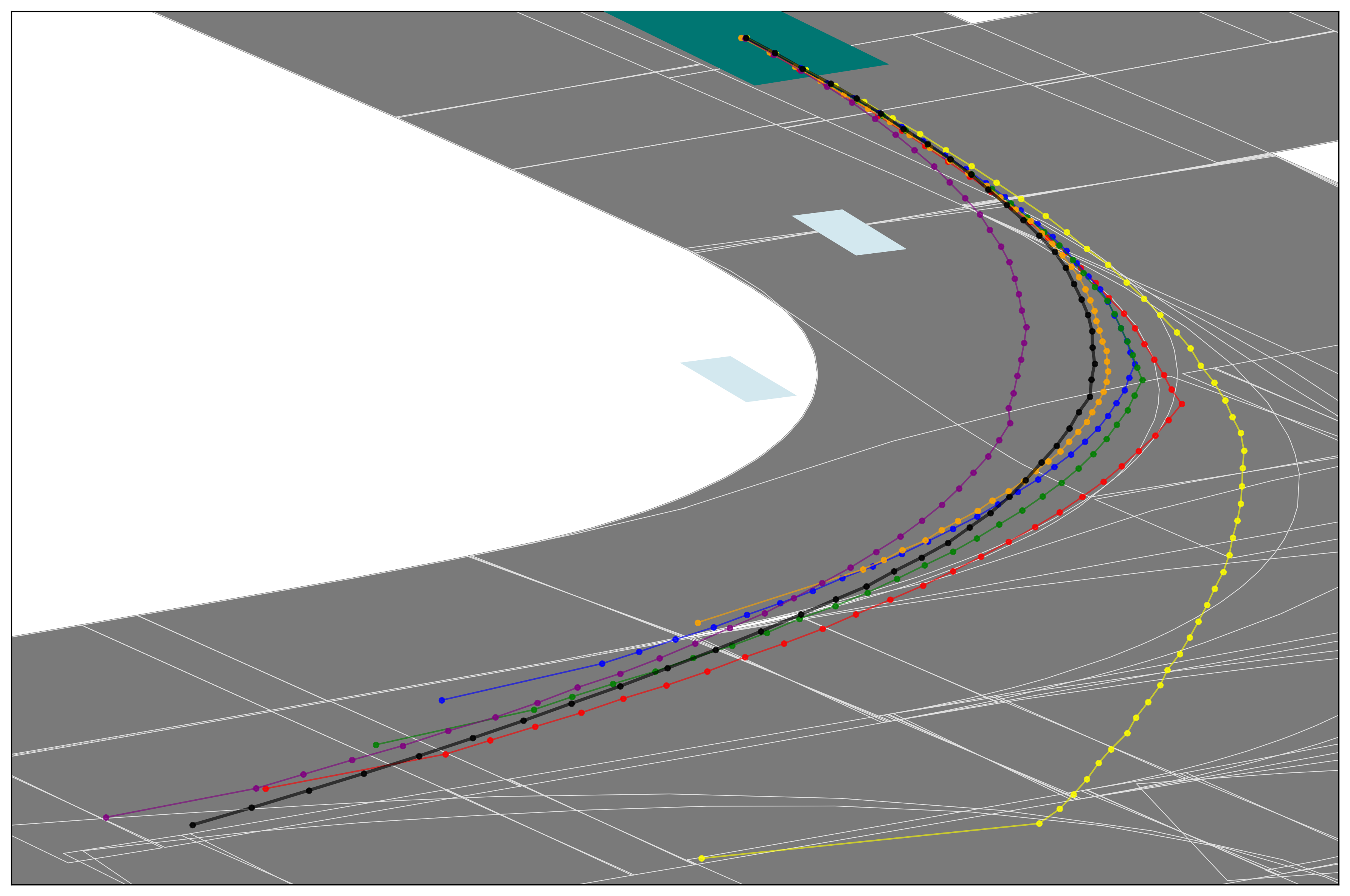}
    \end{subfigure}
    \begin{subfigure}[b]{0.32\textwidth}
        \centering
        \includegraphics[width=\textwidth]{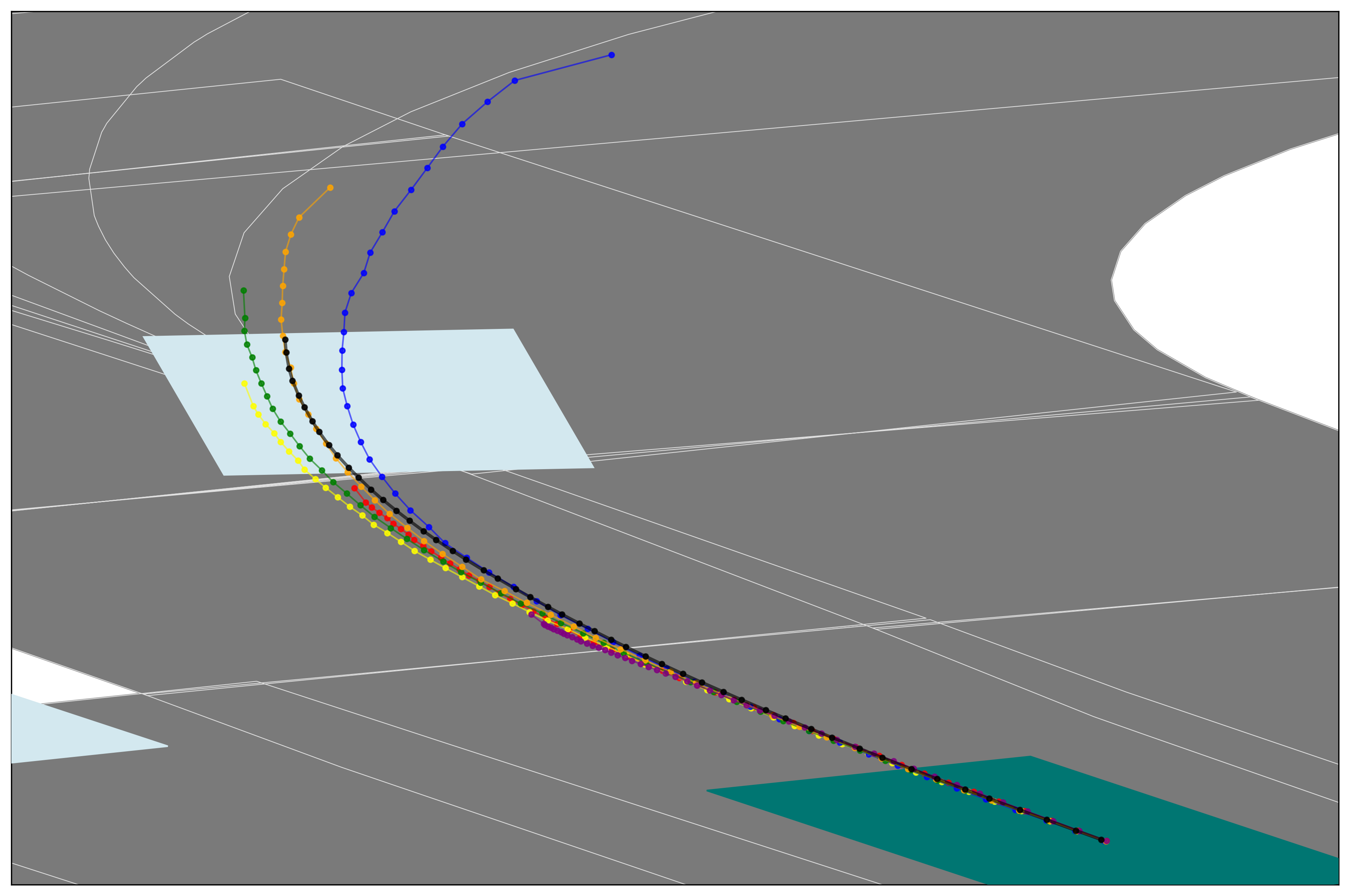}
    \end{subfigure}
    
    \caption{Qualitative results of V2X-RECT in turning scenarios with varying traffic densities from the V2X-Seq dataset. The target vehicle is shown in orange, the predicted vehicles are described as green and other vehicles are depicted in gray. Non-motorized road users are represented as gray dots. Ground-truth trajectories are drawn in black, and predicted trajectories are shown in various colors to distinguish different prediction modes.}
    \label{fig3}
\end{figure*}

\begin{figure*}[h]
    \centering
    \begin{subfigure}[b]{0.32\textwidth}
        \centering
        \includegraphics[width=\textwidth]{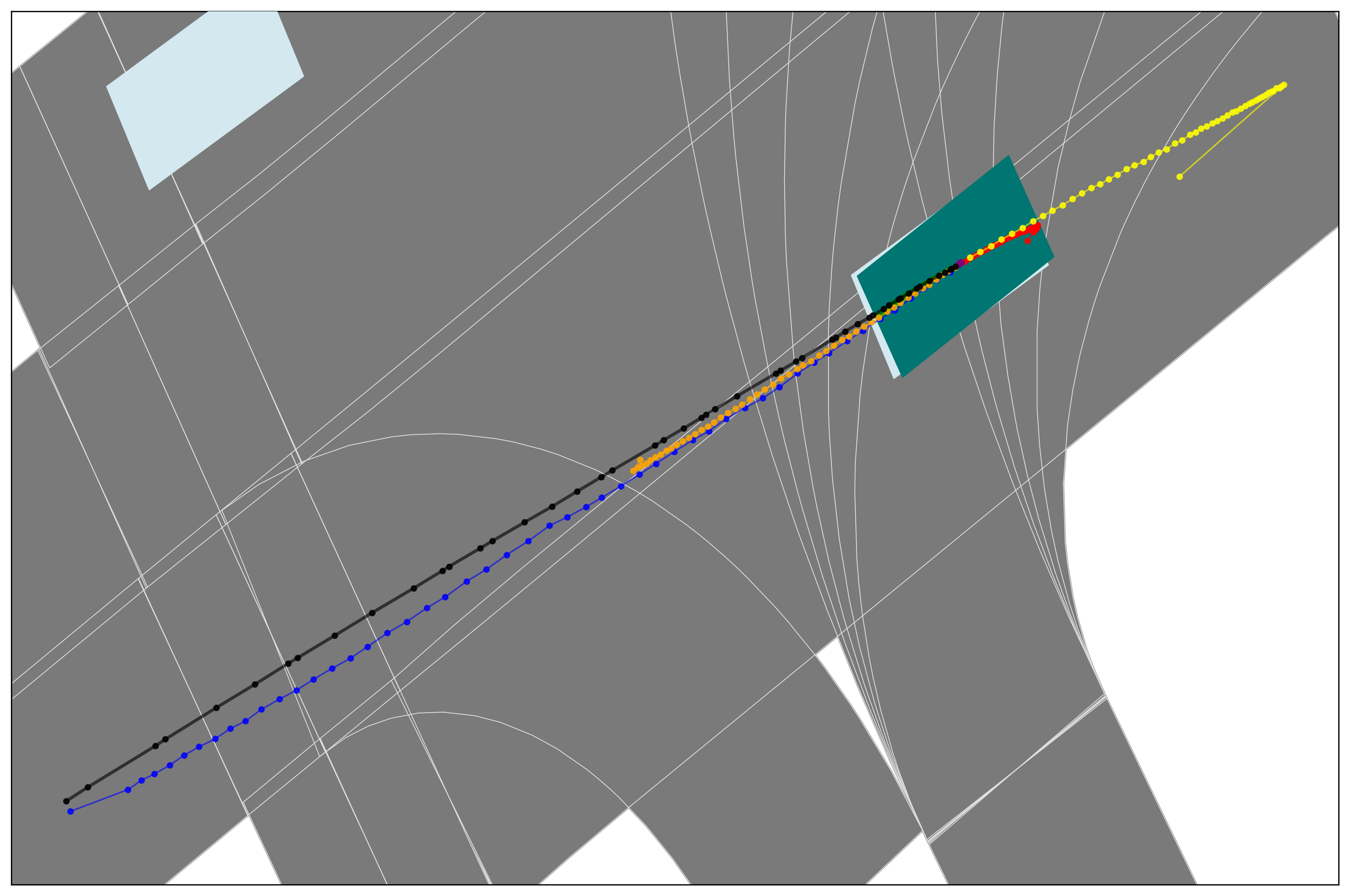}
    \end{subfigure}
    \begin{subfigure}[b]{0.32\textwidth}
        \centering
        \includegraphics[width=\textwidth]{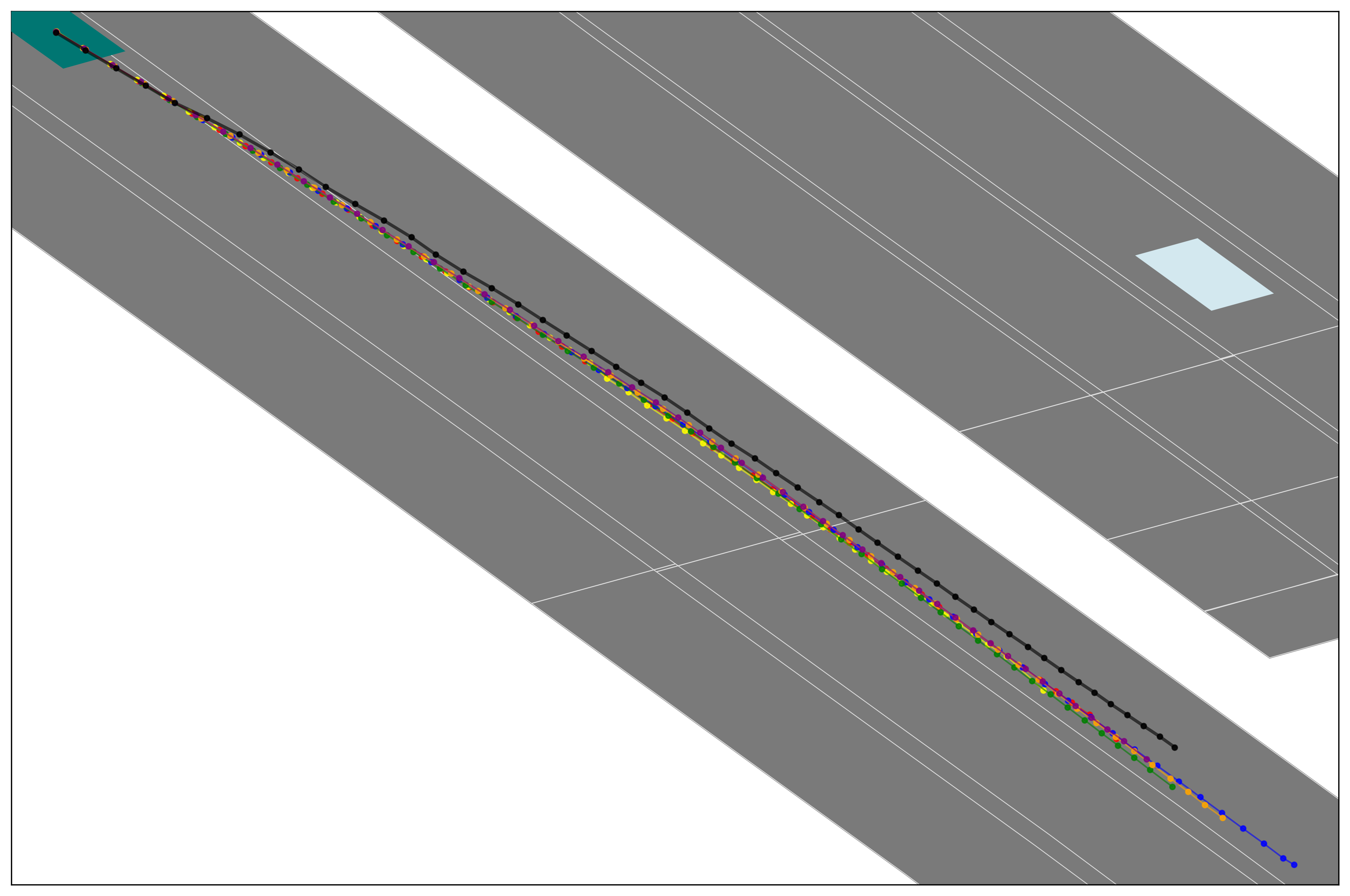}
    \end{subfigure}
    \begin{subfigure}[b]{0.32\textwidth}
        \centering
        \includegraphics[width=\textwidth]{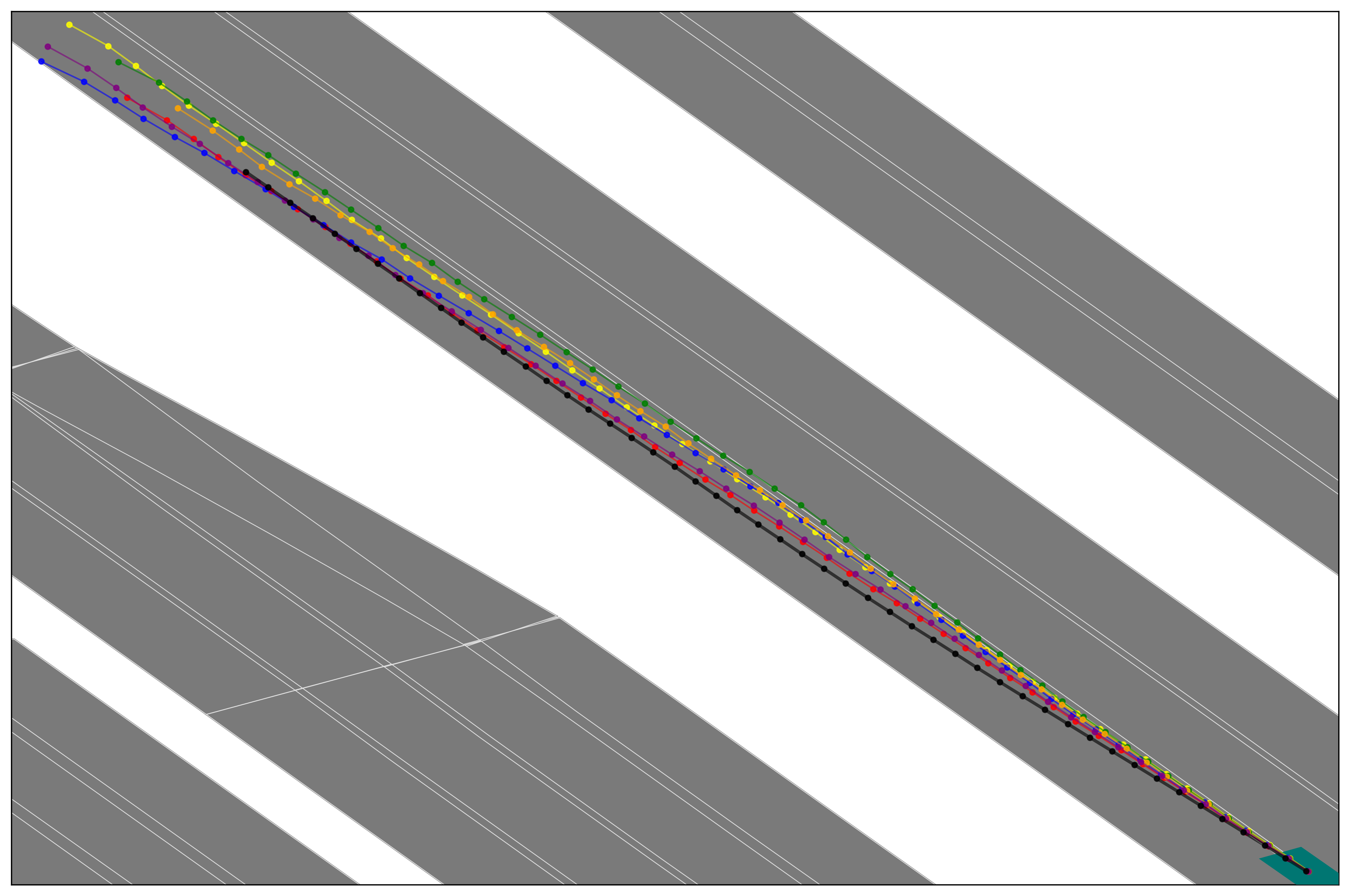}
    \end{subfigure}


    \begin{subfigure}[b]{0.32\textwidth}
        \centering
        \includegraphics[width=\textwidth]{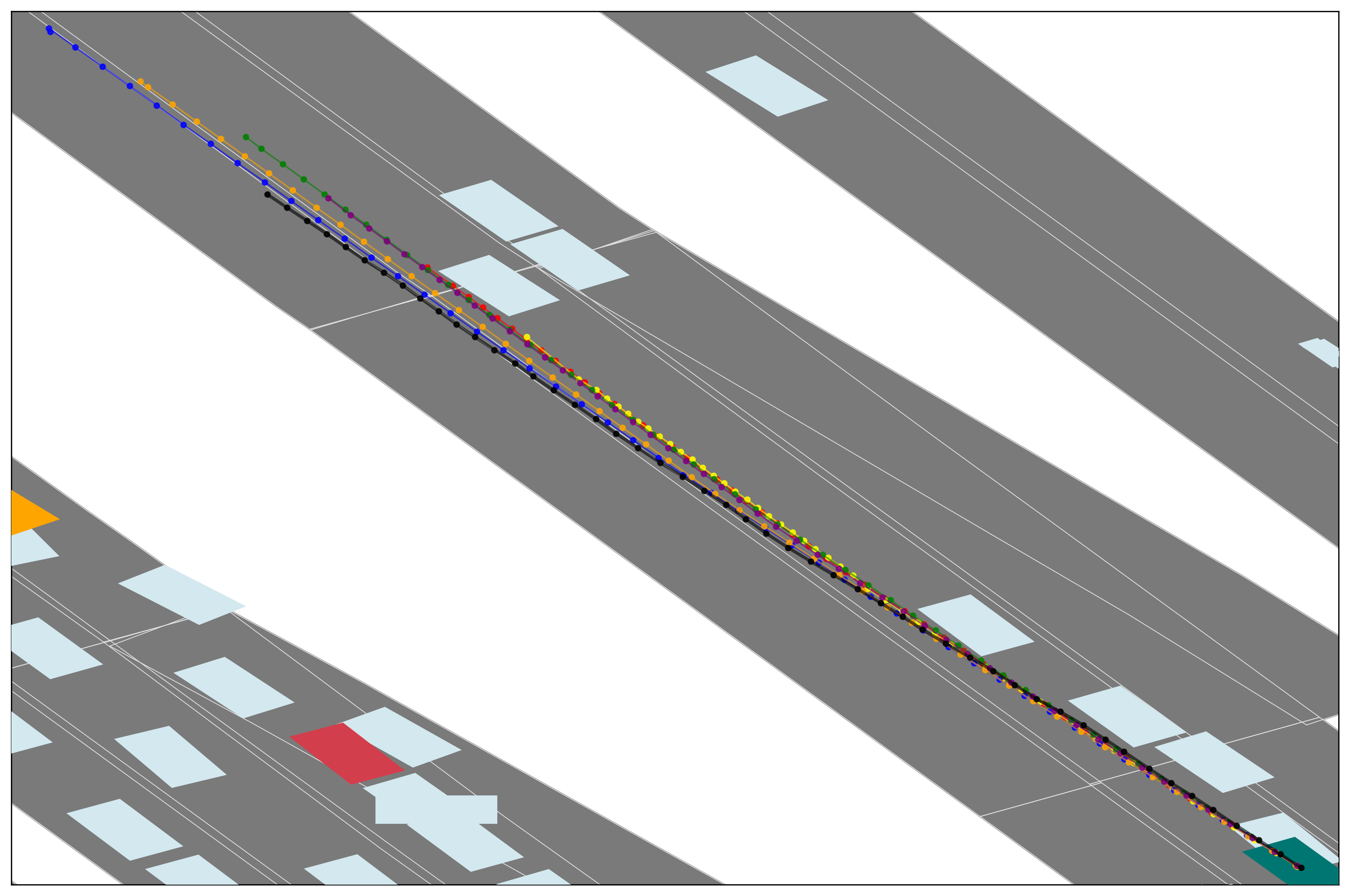}
    \end{subfigure}
    \begin{subfigure}[b]{0.32\textwidth}
        \centering
        \includegraphics[width=\textwidth]{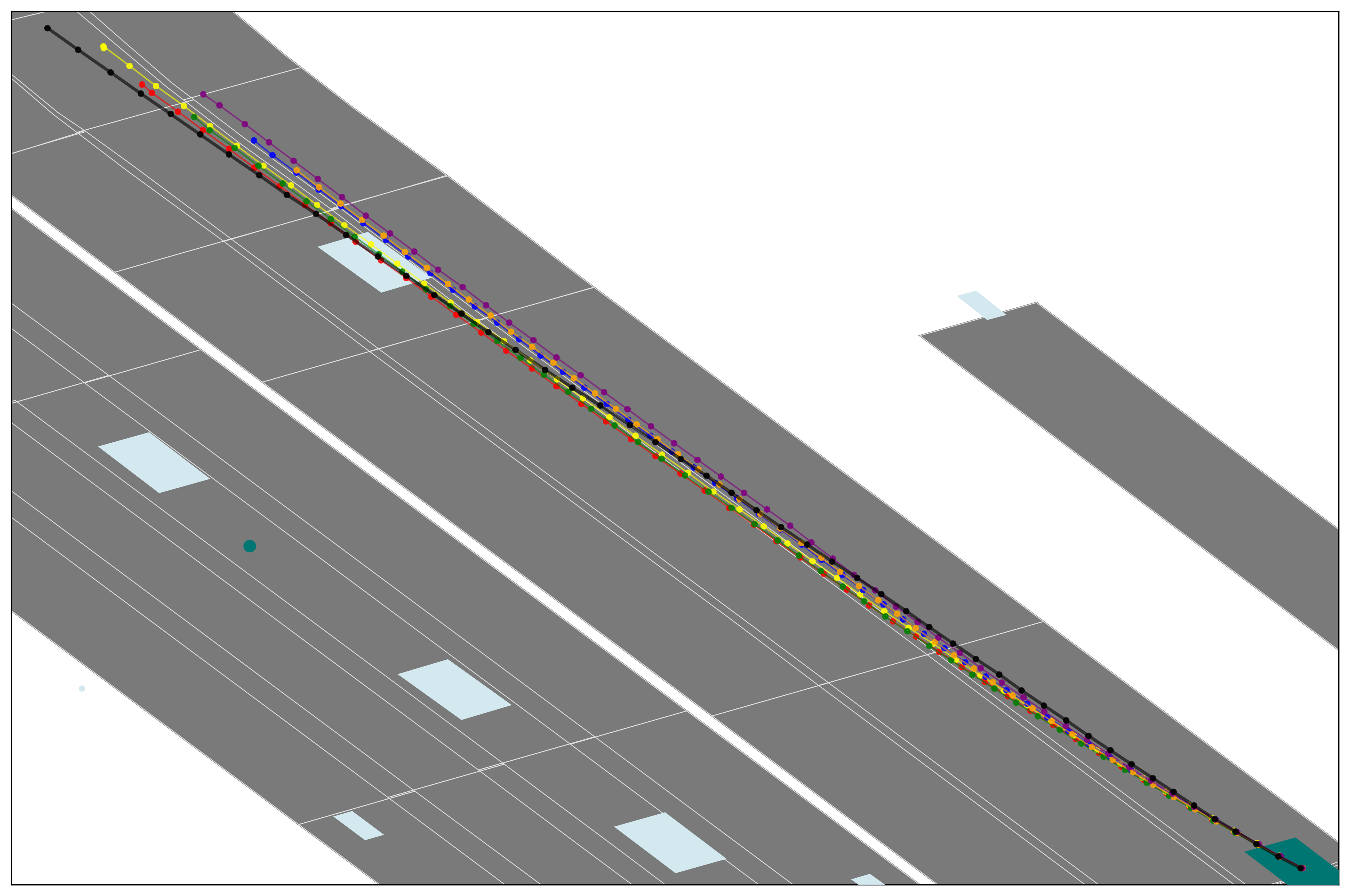}
    \end{subfigure} 
    \caption{Qualitative results of V2X-RECT in lane-changing scenarios with varying traffic densities from the V2X-Seq dataset. The target vehicle is shown in orange, the predicted vehicles are described as green and other vehicles are depicted in gray. Non-motorized road users are represented as gray dots. Ground-truth trajectories are drawn in black, and predicted trajectories are shown in various colors to distinguish different prediction modes.}
    \label{fig4}
\end{figure*}

\begin{figure*}[h]
    \centering
    \begin{subfigure}[b]{0.32\textwidth}
        \centering
        \includegraphics[width=\textwidth]{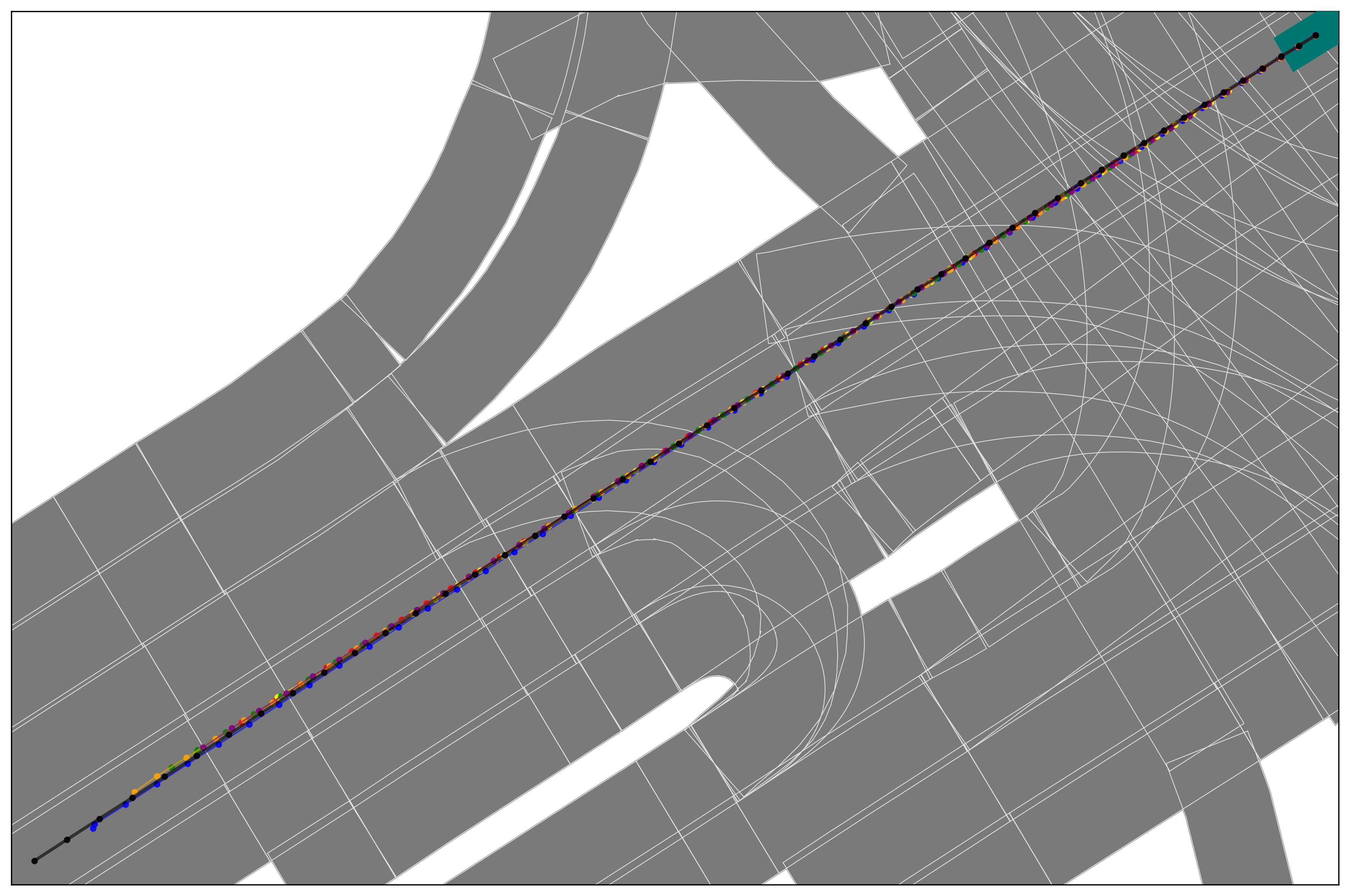}
    \end{subfigure}
    \begin{subfigure}[b]{0.32\textwidth}
        \centering
        \includegraphics[width=\textwidth]{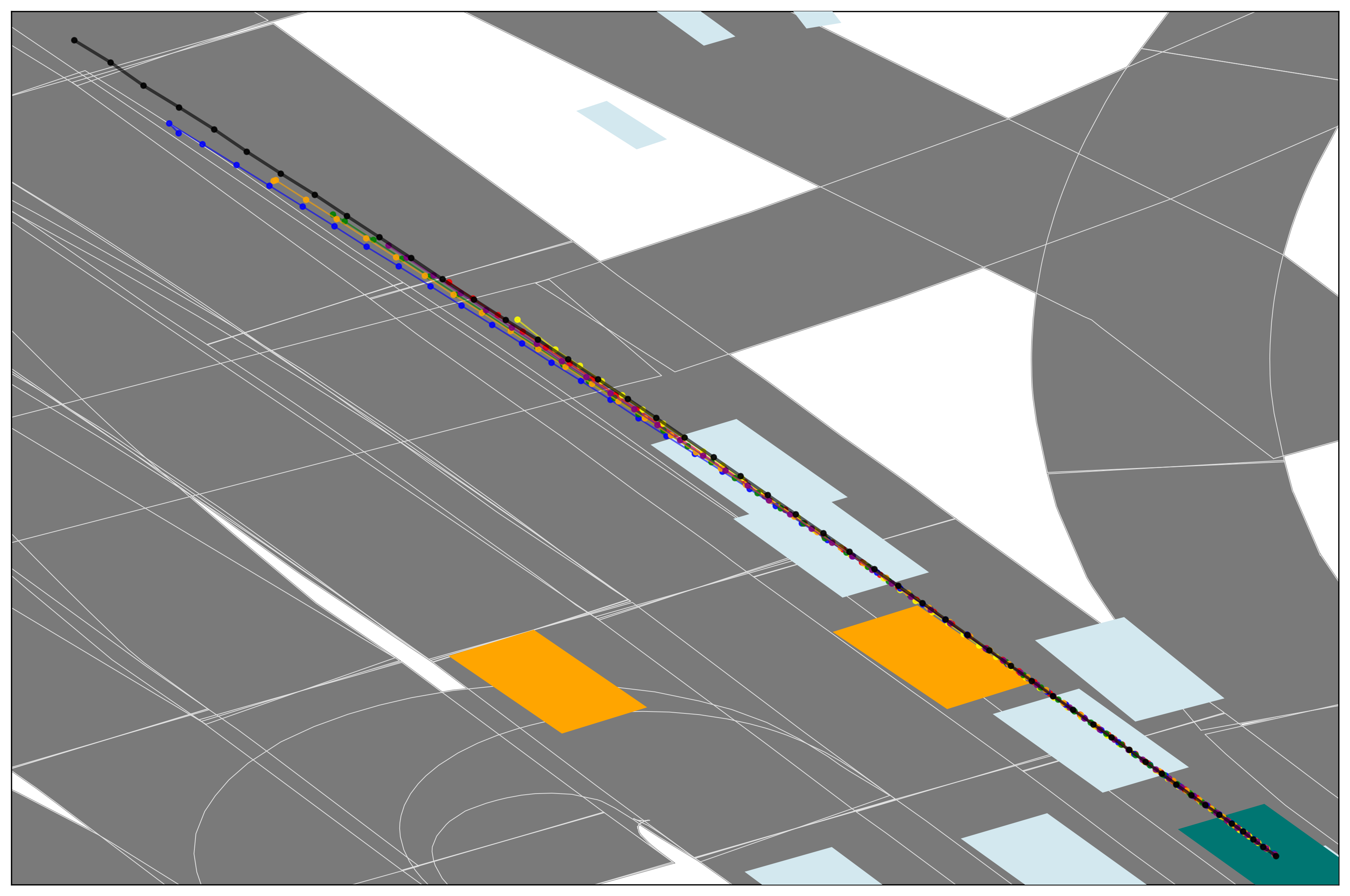}
    \end{subfigure}
    \begin{subfigure}[b]{0.32\textwidth}
        \centering
        \includegraphics[width=\textwidth]{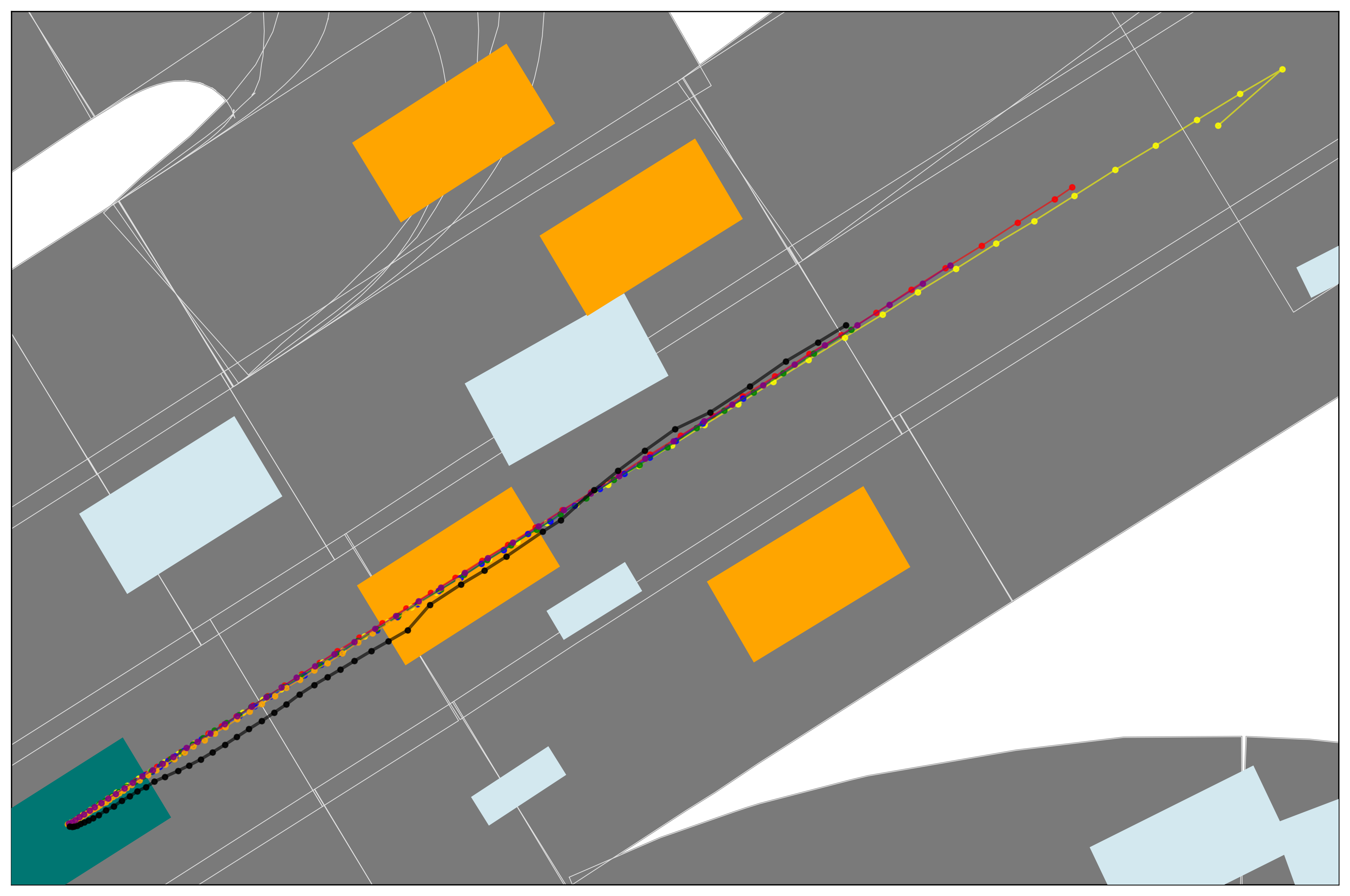}
    \end{subfigure}
    \begin{subfigure}[b]{0.32\textwidth}
        \centering
        \includegraphics[width=\textwidth]{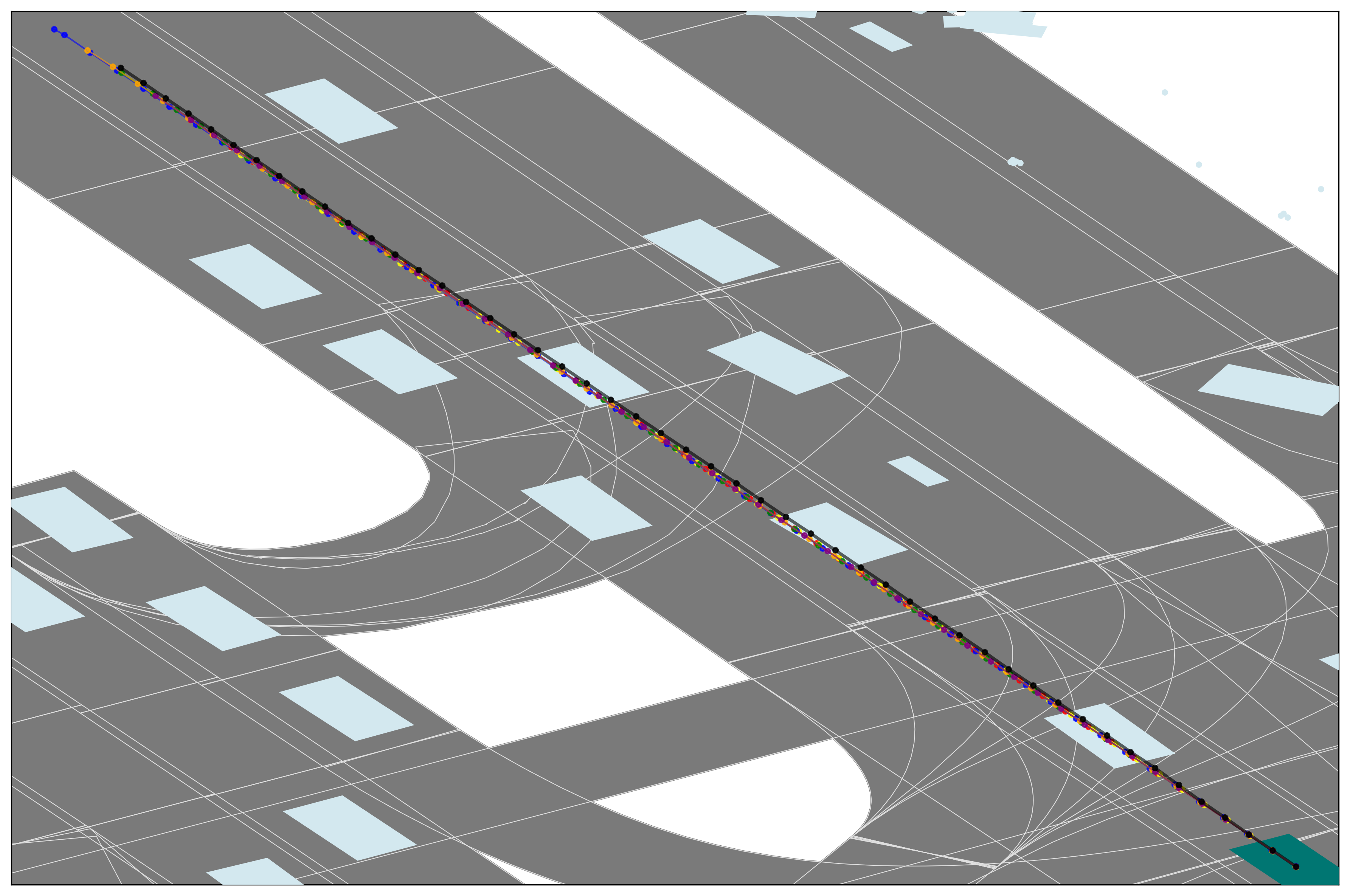}
    \end{subfigure}
    \begin{subfigure}[b]{0.32\textwidth}
        \centering
        \includegraphics[width=\textwidth]{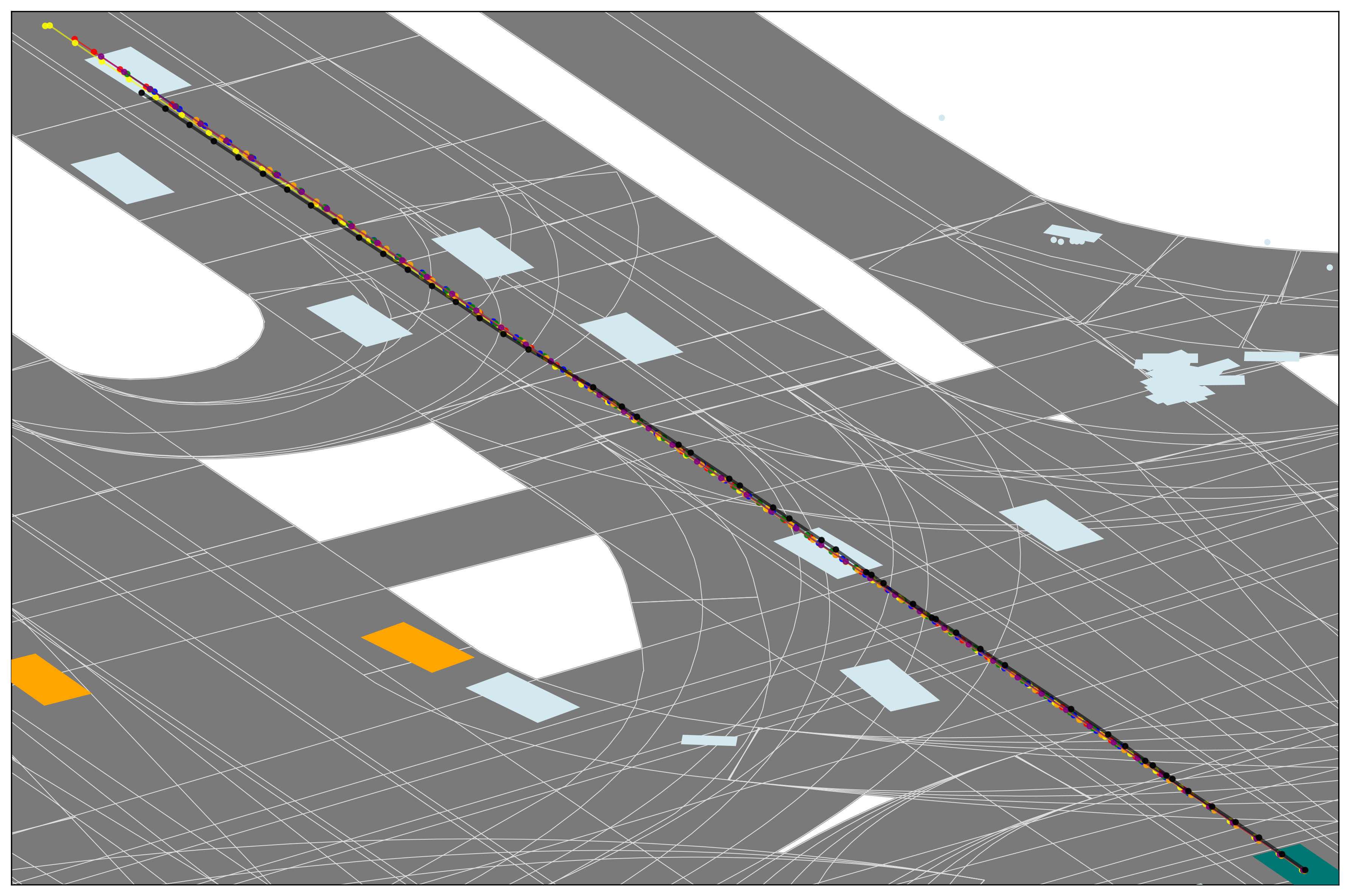}
    \end{subfigure}
    
    \caption{Qualitative results of V2X-RECT in going straight scenarios with varying traffic densities from the V2X-Seq dataset. The target vehicle is shown in orange, the predicted vehicles are described as green and other vehicles are depicted in gray. Non-motorized road users are represented as gray dots. Ground-truth trajectories are drawn in black, and predicted trajectories are shown in various colors to distinguish different prediction modes.}
    \label{fig5}
\vspace{-0.2em}
\end{figure*}

\subsubsection{Comparison With Baselines Across Density Levels}
To evaluate the robustness of our proposed V2X-RECT in complex environments, we compare with V2INet under different traffic density levels on the V2X-Seq dataset. As shown in Table~\ref{table5}, traffic scenes are categorized into six groups according to the number of agents. For each group, we report three standard evaluation metrics: minADE, minFDE, and MR. Across all density groups, V2X-RECT consistently outperforms V2INet in all three metrics. Notably, when the number of agents is below 490, V2INet's performance degrades significantly as the number of agents increases, while V2X-RECT maintains a remarkably stable performance. This phenomenon can be attributed to the design of our model, which focuses on selecting key interaction relationships among agents while efficiently filtering out redundant interactions in high-density traffic environments. This design enhances the robustness of V2X-RECT under varying traffic densities. However, it is noteworthy that when the number of agents exceeds 490, a slight degradation occurs in minADE and minFDE. This abnormality may be closely related to agent behavior patterns in high-density traffic scenarios. As traffic density increases, agents tend to reduce more aggressive strategies, such as lane changes and overtaking maneuvers. Moreover, the decrease in the overall average speed of traffic flow is often associated with more stable behavioral patterns, which in turn reduce the complexity of accurate trajectory prediction.

\begin{table}[t!]
\footnotesize
\caption{Comparisons of the number of parameters.}
\label{table6}
\centering
\begin{tabular}{>{\centering\arraybackslash}m{3.2cm}>{\centering\arraybackslash}m{2.2cm}}
\toprule
Method & Model Parameters\\
\midrule
HIVT\_PP-VIC\cite{refe29*}(2024*) & 2.6M\\
DenseTNT\cite{refe10*} & \textbf{1.0M}\\
V2INet\cite{refe11*} & 26.4M\\
V2X-Graph\cite{refe13*} & 5.0M\\
I2XTraj\cite{refe2} & 3.18M\\
V2X-RECT(Ours)  & \underline{1.3M}\\
\bottomrule
\end{tabular}
\vspace{-0.3em}
\end{table}

\subsection{Qualitative Results}
\subsubsection{Model Complexity and Inference Efficiency Analysis}
To comprehensively evaluate the model's complexity and inference efficiency, we first compared the number of parameters of V2X-RECT with five representative advanced methods: HiVT, DenseTNT, V2INet, V2X-Graph, and I2XTraj. As shown in Table \ref{table6}, the parameter size of V2X-RECT is 1.3M, which outperforms HiVT (2.6M), V2INet (26.4M), I2XTraj (3.18M), and V2X-Graph (5.0M), and is only slightly larger than DenseTNT (1.0M). Among them, DenseTNT and HiVT are single-view trajectory prediction methods. This indicates that V2X-RECT achieves a lightweight model structure while maintaining high prediction performance, making it more suitable for deployment on vehicle-end devices with limited computing resources.\\
\indent To evaluate the inference efficiency of different models, we compared the runtime performance of V2X-RECT against five representative state-of-the-art methods, namely HiVT, DenseTNT, V2INet, V2X-Graph and I2XTraj on a single NVIDIA A100 GPU. As shown in Table\ref{table7}, V2X-RECT achieved an average inference time of 0.104386 seconds, significantly outperforming HIVT\_PP-VIC (0.852635 seconds) and V2INet (0.346285 seconds). Specifically, MV2X-Net reduced the inference latency by approximately 87.8\% compared to HIVT\_PP-VIC and 69.8\% compared to V2INet. This substantial improvement demonstrates the superior computational efficiency of our proposed method. Thanks to the reuse of historical trajectory encoding and the parallelized design, MV2X-Net was more suitable for real-time deployment in resource-constrained V2X scenarios.

\begin{table}[t!]
\footnotesize
\caption{Comparison of Average Inference Time on the Validation Set (in milliseconds).}
\label{table7}
\centering
\begin{tabular}{>{\centering\arraybackslash}m{3.2cm}>{\centering\arraybackslash}m{2.2cm}}
\toprule
Method & Inference Time\\
\midrule
HIVT\_PP-VIC\cite{refe29*}(2024*) & 852.64\\
V2INet\cite{refe11*}(2024*) & \underline{346.29}\\
V2X-RECT(Ours)  & \textbf{104.38}\\
\bottomrule
\end{tabular}
\end{table}

As shown in Figs.~\ref{fig3}, \ref{fig4}, and \ref{fig5}, we visualize the predicted trajectories for three typical traffic environments: turning, going straight, and lane changing, and we present the performance under different traffic densities. In these cases, red solid lines represent the observed trajectory, and the yellow dashed paths and blue solid paths respectively denote the predicted trajectories and ground truth. From these visual trajectories, we observe that most of the ground truth paths are covered by the predicted 10 future trajectories.\\
\indent These visualized results for the turning scenarios (Fig.~\ref{fig3}) demonstrate that our method can effectively predict reasonable trajectories and flexibly respond to surrounding agents on both sparse and dense scenarios. However, the prediction accuracy in the second and third figures is inferior than that of the other figures. This may be due to the small turning angle and the fact that the vehicle does not follow the lane strictly during the turn, resulting in poor performance of our model, which relies on map and destination/task-driven predictions, struggles with such trajectories.\\
\indent The visualization results for the lane-changing scenarios are similar to turning, as illustrated in Fig.~\ref{fig4}. Our method can effectively predict reasonable trajectories in both sparse and dense scenarios. In the fourth and fifth figures, the prediction results exhibit trajectory multi-modality. As the number of agents in the scene increases, the target agent must account for interactions with surrounding agents during its movement. However, since our method does not model future agent interactions, the prediction performance in high-density scenarios is worse than in low-density ones, although the model still exhibits multi-modal characteristics.\\
\indent The visualization results for the going straight scenarios are shown in Fig.~\ref{fig5}. The prediction performance in the third figure is lower compared to the other figures. This is because the agent encounters a red light while crossing the intersection, leading to multi-modal behavior in our predictions. Specifically, some predictions show the agent stopping, while others predict the agent continuing straight through the intersection. Although our method incorporates traffic light information, the performance of our algorithm has not reached its optimal level due to the scarcity of such data in the dataset.

\subsection{Ablation Study}

We conduct the ablation studies on the main components of V2X-RECT following the same routine as the main experiment. The model performance is evaluated on the V2X-Seq test set. The configurations of six different models are detailed in Table~\ref{table8},~\ref{table9},~\ref{table10}. These components are Signal-Guided Temporal Attention Module(ST-A), Map-Aware Attention Module(M-A), Signal-Guided Social Attention Module(SS-A), Muti-View Correction Module(MVCM), and Feature Alignment Module(FAM). Based on the correlation between modules, we group STA, MA, and SSA into one category, and MVCM and FAM into another category for separate validation. STA first applies Fourier transform to extract the frequency-domain features of agents'historical trajectories. Then, based on traffic signal features and historical trajectory features, it utilizes an attention module to extract trajectory features that vary with traffic signals. The ablated version of the model without STA only employs Fourier transform to extract historical trajectory information. MA first extracts map features, and then uses an attention to fuse map features with agents'historical trajectory features, extracting the location information of agents on the map. The ablated version of the model without MA does not consider the relationship between the map and the agents during the encoding and uses attention to extract relevant features during the decoding. SSA represents the social attention mechanism that, within a certain range, fuses the relationships between the target agent and its surrounding agents through attention to obtain social attention. The ablated version of the model without SSA does not consider interaction relationships during the encoding and uses attention to extract relevant features during the decoding. MVCM corrects ID Switch through mapping relationships of trajectory among multi-views, while FAM represents the feature fusion module based on these mappings. The ablated version of the model without FAM directly fuse trajectory features without map. Tables~\ref{table8},~\ref{table9},~\ref{table10} present the results of the ablation experiments, with the final row showing our results.\looseness=-1

\begin{table}[!t]
\footnotesize
\caption{Ablation study of the signal-guided behavior-interaction module. Experiments is performed on the V2X-Seq validation set.}
\label{table8}
\centering
\begin{tabular}{ccccccc}
\toprule
Method  & ST-A & M-A & SS-A & minADE & minFDE & MR\\
\midrule
MSA& & \ding{51} & \ding{51} & 0.68 & 1.04 & 0.13\\
TSA& \ding{51} &  & \ding{51} & 0.73 & 1.19 & 0.13\\
TMA& \ding{51} & \ding{51}& & 0.64 & 0.92 & 0.12  \\
TMSA& \ding{51} & \ding{51} & \ding{51} & 0.53 & 0.80 & 0.09\\
\bottomrule
\end{tabular}
\vspace{-0.1em}
\end{table}

\begin{table}[t!]
\footnotesize
\caption{Ablation study of the continuous signal-informed mechanism.}
\label{table9}
\centering
\begin{tabular}{cccc}
\toprule
Method & minADE & minFDE & MR\\
\midrule
Remove signals & 0.58& 0.85& 0.10\\
V2X-RECT(Ours) & 0.53 & 0.80 & 0.09\\
\bottomrule
\end{tabular}
\vspace{-0.5em}
\end{table}

\begin{table}[t!]
\footnotesize
\caption{Ablation study of our introduced Multi-View Trajectory Correction and Fusion Module.}
\label{table10}
\centering
\begin{tabular}{cccccc}
\midrule
Method & MVCM & FAM & minADE & minFDE & MR\\
\hline
FM  & & \ding{51} & 0.58 & 0.83 & 0.10\\
CM & \ding{51} &  & 0.54 & 0.84 & 0.10\\
CFM & \ding{51} & & 0.53 & 0.80 & 0.09\\
\bottomrule
\end{tabular}
\vspace{-0.5em}
\end{table}

\noindent \textbf{Importance of The Signal-Guided Temporal Attention Module (ST-A):} To evaluate the significance of capturing agent behavior patterns influenced by traffic signal changes, we conducted comparative experiments between the TMSA and MSA models. TMSA adopts the ST-Attention module to capture the temporal attention of agents, whereas MSA relies on Fourier transform to extract historical features, resulting in lower feature extraction quality. As shown in Table~\ref{table8}, TMSA outperforms MSA across all performance metrics, with improvements of 0.15, 0.24, and 0.04 in minADE, minFDE, and MR, respectively. These results highlight that the ST-A module effectively captures motion pattern transitions induced by dynamic signal changes, significantly enhancing the model's ability to represent time-varying behaviors.

\noindent \textbf{Influence of Map-Aware Attention Module (M-A):} A series of studies are performed to further validate the impact of the MA model. Specifically, TMSA employed the MA model in encoding as the map feature extraction method, while TSA is not adopted this model, both methods integrate map features during the decoding. Removing this module led to a notable performance degradation, with minADE, minFDE, and MR increasing by 0.15, 0.24 and 0.04, respectively. This observation impressively demonstrates that the MA module enhances the model's spatial understanding of environmental and its awareness of the spatial position of the agent by integrating road topology information from high-precision maps.

\noindent \textbf{Effect of The Signal-Guided Social Attention Module (SS-A):} To evaluate the significance of capturing social interaction patterns among agents influenced by traffic signal changes, we conducted comparative experiments between the TMSA and TMA models. TMSA adopts the SS-A module to capture the social attention of agents.  In contrast, TMA omits this module in encoding, both methods integrate social interaction features during the decoding. As shown in Table~\ref{table8}, TMSA outperforms TMA across all performance metrics, with improvements of 0.11, 0.12, and 0.03 in minADE, minFDE, and MR, respectively. Experiments demonstrate that SS-A can effectively utilize traffic signal states to filter out irrelevant interactions. Meanwhile, as shown in Table ~\ref{table9}, we also compare the complete model with a variant that removes the traffic signal encoding module to verify the effectiveness of the continuous signal guidance mechanism. After removing the traffic light information, minADE, minFDE, and MR increased by 0.05, 0.05, and 0.01, respectively. The continuous signal encoding enables the model to more accurately capture changes in signal phases, thereby achieving reliable motion strategies and social interaction.

\noindent \textbf{Influence of The Muti-View Correction Module (MVCM):} To evaluate the impact of MVCM, we conducted comparative experiments between the CFM and FM models. Specifically, CFM employed the MVCM model in encoding as the map feature extraction method, while FM is not adopted this model, both methods integrate map features during the decoding. CFM employed the MVCM model that performs frame-level analysis of the spatio-temporal overlap information of agents across multiple viewpoints, establishes identity matching relationships, and addresses identity switch issues, while FM is not adopted this model. As shown in Table~\ref{table10}, CFM outperforms FM significantly across all performance metrics, with improvements of 0.05, 0.03, and 0.01 in minADE, minFDE, and MR, respectively. The experimental results demonstrate that CFM plays a critical role in enhancing trajectory consistency and prediction reliability.

\noindent \textbf{Impact of the Feature Alignment Module (FAM):} To evaluate the impact of FAM, we conducted comparative experiments between the CFM and CM models. Specifically, CFM employs the FAM module to align features between agents and then uses cross-attention to fuse information from different sources, whereas CM directly fuses the information without alignment. As shown in Table~\ref{table10}, CFM outperforms CM significantly across all the performance metrics, with improvements of 0.01, 0.04, and 0.01 in minADE, minFDE, and MR, respectively. The experimental results demonstrate that our CFM employed its alignment module to effectively support multi-view trajectory fusion, enabling more efficient cross-view attention-based integration.

\section{Conclusion and Discussions}
\label{sec6}

In this paper, we present a trajectory prediction framework specifically designed for high-density V2X scenarios. First, the identity matching and correction module is designed for V2X scenarios, aiming to address issues of ID switch and achieve identity matching across multi-view agents. Second, a traffic signal-guided interaction and behavior modeling module is introduceed to identify effective interactions among agents controled by the same traffic signal and capture the dynamic influence of continuous signal changes on agent behavior. Finally, a contextual feature reuse method based on a local spatio-temporal coordinate system is proposed to enable parallel decoding and accurately predict future trajectories for multiple agents. Extensive experimental results on the mainstream V2X-Seq and V2X-Traj datasets demonstrate that our V2X-RECT outperforms existing state-of-the-art methods by 30\%-50\% on key metrics such as minADE, minFDE, and MR with enhanced robustness and inference efficiency across different traffic densities.

In the future, we plan to tackle the issue of temporal misalignment across different viewpoints and investigate strategies for coordinating information resources across vehicles and roadside units. This will help reduce information redundancy and communication overhead while enabling the exploitation of diverse agent appearance features, further enhancing the intelligence and real-time capabilities of the system.

\end{document}